\documentclass{article}
\usepackage[utf8]{inputenc}
\usepackage[a4paper, margin=3cm]{geometry}

\usepackage[english]{babel}

\usepackage[backend=biber,
style=apa,
maxcitenames=2,
maxbibnames=100, 
language=english,
doi=false,
isbn=false,
url=false,
uniquename=false
]{biblatex}
\DeclareLanguageMapping{english}{english-apa} 
\usepackage{subcaption}
\usepackage{placeins}
\usepackage{mathtools}
\usepackage{amssymb}
\usepackage{amsmath}
\usepackage{amsfonts} 
\usepackage{bbm}
\usepackage{fancyhdr}
\usepackage{times}
\usepackage{amsthm}
\usepackage{siunitx}
\usepackage{nicefrac}
\usepackage{hyperref}
\usepackage{url} 
\usepackage{calc}
\usepackage{cleveref}

\hypersetup{
    colorlinks=true,
    linkcolor=darkblue,
    filecolor=darkblue,
    urlcolor=darkblue,
    citecolor=darkblue,
    pdfauthor={U. K\"{o}the, P. Sorrensen},
    pdftitle={A Review of Change of Variable Formulas for Generative Modeling}
}

\usepackage{algorithm}
\usepackage{algorithmic}

\newcommand{\x}{\boldsymbol{x}}
\newcommand{\X}{\boldsymbol{X}}
\newcommand{\xw}{\widetilde{\boldsymbol{x}}}
\newcommand{\XW}{\widetilde{\boldsymbol{X}}}
\newcommand{\y}{\boldsymbol{y}}
\newcommand{\Y}{\boldsymbol{Y}}
\newcommand{\z}{\boldsymbol{z}}
\newcommand{\Z}{\boldsymbol{Z}}

\renewcommand{\S}{\boldsymbol{S}}
\newcommand{\s}{\boldsymbol{s}}
\newcommand{\J}{\boldsymbol{J}}
\newcommand{\A}{\boldsymbol{A}}
\newcommand{\Q}{\boldsymbol{Q}}
\newcommand{\mub}{\boldsymbol{\mu}}

\newcommand{\sigmab}{\boldsymbol{\sigma}}
\newcommand{\Sigmab}{\boldsymbol{\Sigma}}

\newcommand{\0}{\boldsymbol{0}}
\newcommand{\U}{\boldsymbol{U}}
\newcommand{\V}{\boldsymbol{V}}
\newcommand{\W}{\boldsymbol{W}}
\newcommand{\w}{\boldsymbol{w}}
\newcommand{\norm}[1]{\left\lVert#1\right\rVert}
\newcommand{\Var}{\mathrm{Var}}

\newcommand{\given}{\,|\,}

\newcommand{\kl}{\text{KL}}
\newcommand{\FS}{\mathcal{F}}
\newcommand{\IS}{\mathcal{I}}
\newcommand{\MS}{\mathcal{M}}
\newcommand{\XS}{\mathcal{X}}
\newcommand{\TS}{\mathcal{T}}
\newcommand{\ZS}{\mathcal{Z}}
\newcommand{\normal}{\mathcal{N}}
\newcommand{\eye}{\mathbb{I}}
\newcommand{\E}{\mathbb{E}}
\newcommand{\R}{\mathbb{R}}
\newcommand{\integers}{\mathbb{Z}}

\newlength{\depthofsumsign}
\newcommand{\nsum}[1][1.4]{
    \mathop{%
        \raisebox
            {-#1\depthofsumsign+1\depthofsumsign}
            {\scalebox
                {#1}
                {$\displaystyle\sum$}%
            }
    }
}

\DeclareMathOperator{\diag}{diag}
\DeclareMathOperator{\tr}{tr}

\DeclareMathOperator{\dom}{dom}

\DeclareMathOperator{\slice}{slice}
\DeclareMathOperator{\pad}{pad}

\DeclareMathOperator{\argmin}{\arg\min}

\DeclareMathOperator{\uniform}{uniform}

\usepackage{float}

\usepackage{accents}

\usepackage{enumitem}

\usepackage{xcolor}
\definecolor{darkblue}{RGB}{0,0,128}
\definecolor{darkgreen}{RGB}{0, 128, 0}
\definecolor{darkred}{RGB}{128, 0, 0}
\definecolor{black}{RGB}{0, 0, 0}
\definecolor{errorcolor}{HTML}{481567}
\definecolor{viridisgreen}{HTML}{55C667}
\definecolor{darkgray}{RGB}{70, 70, 70}

\usepackage{yfonts}

\newcommand{\todo}[1]{}
\newcommand{\CoV}[1]{\quad\text{\bf#1:}}

\addto\extrasenglish{

}

\addto\extrasenglish{

}

\addto\extrasenglish{
  
}

\title{A Review of Change of Variable Formulas \\ for Generative Modeling}
\author{Ullrich Köthe}
\date{\normalsize \href{mailto:ullrich.koethe@iwr.uni-heidelberg.de}{ullrich.koethe@iwr.uni-heidelberg.de} \\
      Heidelberg University\\
      Working Paper, August 2023}

\begin{document}

\maketitle

\begin{abstract}
\noindent Change-of-variables (CoV) formulas allow to reduce complicated probability densities to simpler ones by a learned transformation with tractable Jacobian determinant.
They are thus powerful tools for maximum-likelihood learning, Bayesian inference, outlier detection, model selection, etc.
CoV formulas have been derived for a large variety of model types, but this information is scattered over many separate works.
We present a systematic treatment from the unifying perspective of encoder/decoder architectures, which collects 28 \todo{(check)} CoV formulas in a single place, reveals interesting relationships between seemingly diverse methods, emphasizes important distinctions that are not always clear in the literature, and identifies surprising gaps for future research.
\end{abstract}

\section{Introduction}

A change-of-variables (CoV) formula allows to express a complicated probability density $p(\X)$ in terms of a simpler and known one $p(\Z)$, utilizing a suitable transformation between the variables $\X$ and $\Z$.
Change-of-variables formulas have become popular because they play a central role in the training and application of {\em normalizing flows} (NFs).
It is less well known that bottleneck architectures like autoencoders and variational autoencoders, and indeed many other model types, admit similar formulas as well.
Moreover, existing reviews, e.g. \autocite{kobyzev2021normalizing_flows,papamakarios2021normalizing_flows,nielsen2020survae}, only cover change-of-variables formulas for a single model type.
This motivated us to conduct a systematic review across all model architectures, whose results are presented in this paper.

Models admitting a change-of-variables formula are capable of {\em density estimation} -- i.e., they can efficiently calculate the probability density $p(\X\!=\!\x)$ of a given data instance $\x$.
Density estimation is an important enabler for various downstream tasks, for example, Bayesian data inference, outlier detection, comparison of competing hypotheses, or composition into larger statistical models, to name just a few.
It also enables maximum likelihood training, a theoretically appealing method to optimize model performance by minimizing the KL-divergence $\kl[p^*(\X) \,||\, p(\X)]$ between a true and an approximate distribution.

To make sense of the commonalities and differences between change-of-variables formulas, we consider them in the context of {\em generative models} -- i.e., models that not only can calculate densities of given data instances, but also create synthetic data $\x \sim p(\X)$ according to the density of interest.
The two tasks correspond to opposite model execution directions: density estimation transforms data from $\X$-space to $\Z$-space (called {\em encoding}), and generation proceeds the other way around ({\em decoding}).
Accordingly, we refer to the $\Z$ variables as {\em codes}, and their relationship to the data $\X$ can either be deterministic, $\x = g(\z)$, or stochastic, $\x \sim p(\X\given\Z\!=\!\z)$ (see \autoref{sec:basic-concepts} for notation details).

Treating encoding and decoding jointly is beneficial, because consistency between both directions (or the lack thereof) is a powerful tool for the analysis and understanding of model properties.  
Change-of-variables formulas take many different forms for different model architectures, depending on
\begin{itemize}
    \item whether the dimension of $\Z$ is smaller, larger, or equal to the dimension of $\X$,
    \item whether the model is stochastic or deterministic,
    \item whether the transformation is implemented as a finite composition of functions (e.g. layers of a neural network) or as a sequence of infinitesimal steps (e.g. an ordinary differential equation), and
    \item whether or not codes and their relationship to the data are known beforehand (enabling supervised learning) or not (requiring unsupervised learning).
\end{itemize}
A categorization guided by these differences led us to the identification of four fundamental model types:
\begin{description}
    \item[ Bijective flows:] The decoder realizes a deterministic bijective mapping from code to data space, and the encoder is its inverse.
    Data and codes have the same dimension, and the encoding is lossless.
    Invertible neural networks and diffusion models are popular instances of this type (see \autoref{sec:bijective-flows}).
    \item[ Injective flows:] Injective models have a bottleneck architecture -- i.e., the codes have fewer dimensions than the data.
    Encoding is thus generally lossy, unless the data ``live'' in a lower dimensional subspace to begin with.
    Both decoder and encoder are deterministic. 
    The decoder maps codes injectively to a manifold embedded in the data space, and the encoder is (or, in practice, approximates) its surjective pseudo-inverse.
    Autoencoders are the prime example of this architecture (\autoref{sec:injective-flows}). The term ``injective flows'' was coined by \textcite{kumar2020regularized}.
    \item[ Split flows:] This model type combines the properties of bijective and injective flows and supports both lossless and lossy encoding.
    In lossless mode, a split flow works bijectively, but splits the code space into a ``core'' and a ``detail'' part.
    The former encodes the essential properties of the data, and the latter the deviations from the core behavior.
    In lossy mode, one ignores the encoder's detail output and considers only the core dimensions as codes.
    The decoder maps these core codes onto a manifold embedded in the data space, but additionally samples suitable off-manifold deviations to reconstruct the full data distribution.
    Thus, split flows combine a deterministic encoder with a stochastic decoder.
    This model type has received relatively little attention in the context of deep learning, with denoising normalizing flows \autocite[see \autoref{sec:split-nfs}]{horvat2021denoising} and general incompressible flows \autocite[GIN,][see \autoref{sec:finite-nfs}]{sorrenson2019disentanglement} being two notable examples we are aware of.
    \item[ Stochastic flows:] Here, both encoder and decoder are stochastic and realize conditional probabilities $p(\Z\given\X)$ and $p(\X\given\Z)$ respectively, instead of the functions $\z=f(\x)$ and $\x=g(\z)$.
    Consequently, there is no unique code for a given data instance and no unique data point for a given code, and the encoding is always lossy.
    This model type may employ a bottleneck (as in variational autoencoders, \autoref{sec:bayesian-models}) or preserve the data dimension (as in diffusion by stochastic differential equations, \autoref{sec:stochastic-differential-equations}).
\end{description}
Interestingly, these types represent alternative resolutions of a triple trade-off that has been identified by modern coding theory, namely between coding rate, reconstruction error, and perceptual quality of the generated data.
We will discuss this connection in \autoref{sec:coding-theory}.

\begin{figure}[t]
    \centering
    \begin{subfigure}{.3\textwidth}
      \centering
      \includegraphics[width=.9\linewidth]{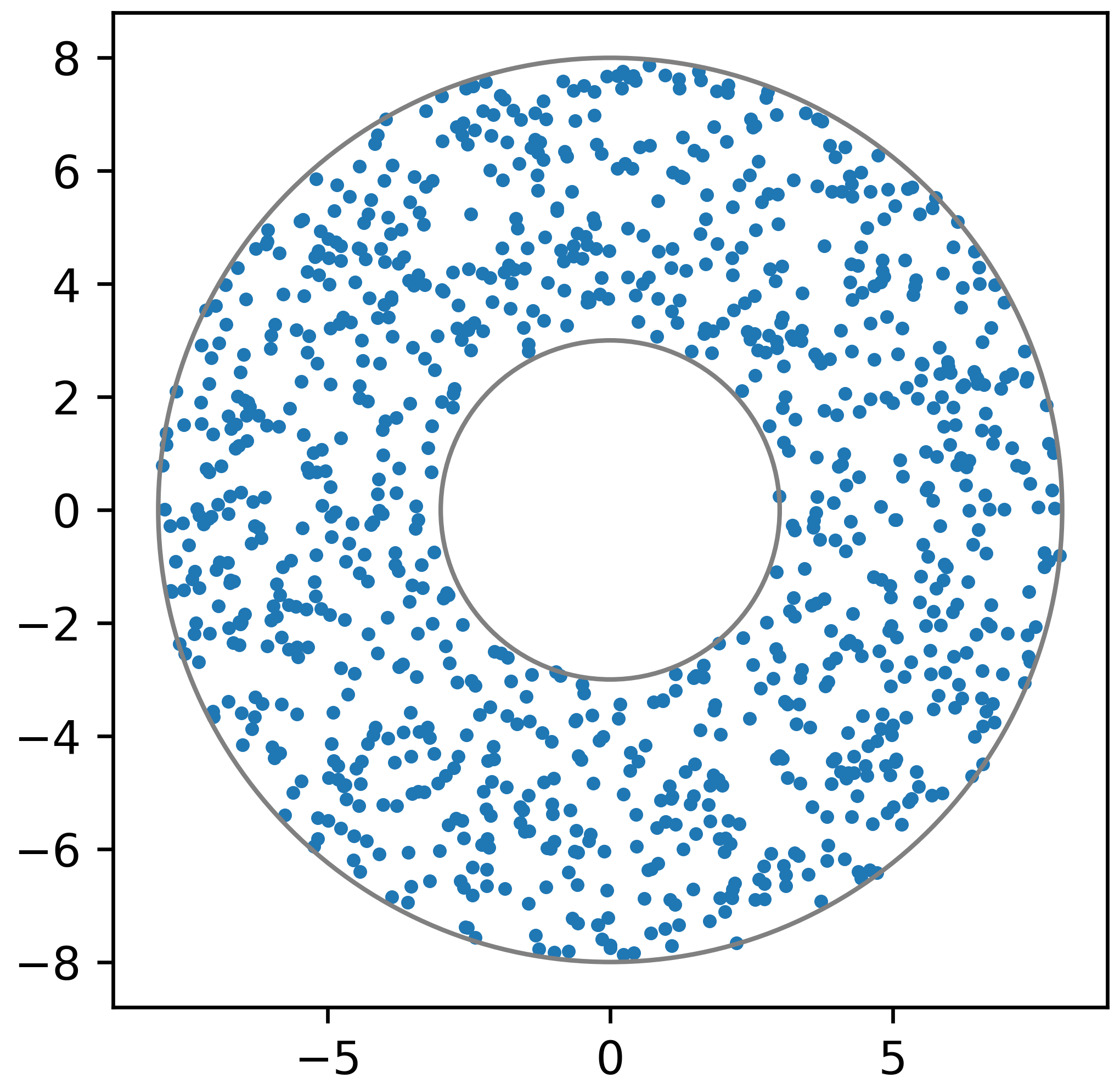}
      \caption{Target distribution}
      \label{fig:donut-target}
    \end{subfigure}%
    \begin{subfigure}{.3\textwidth}
      \centering
      \includegraphics[width=.92\linewidth]{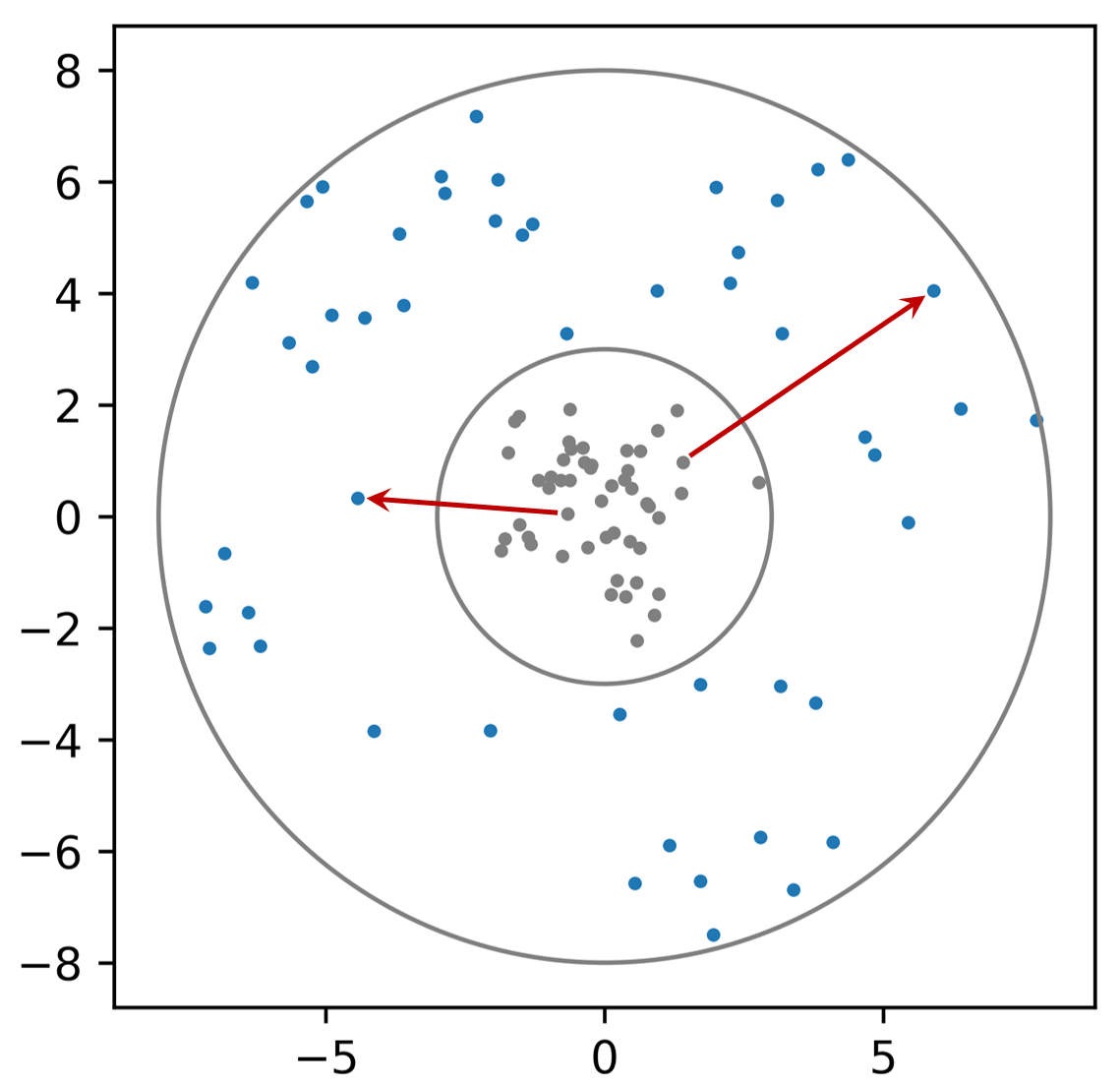}
      \caption{Bijective flow representation}
      \label{fig:donut-nf}
    \end{subfigure}%
    \begin{subfigure}{.3\textwidth}
      \centering
      \includegraphics[width=.9\linewidth]{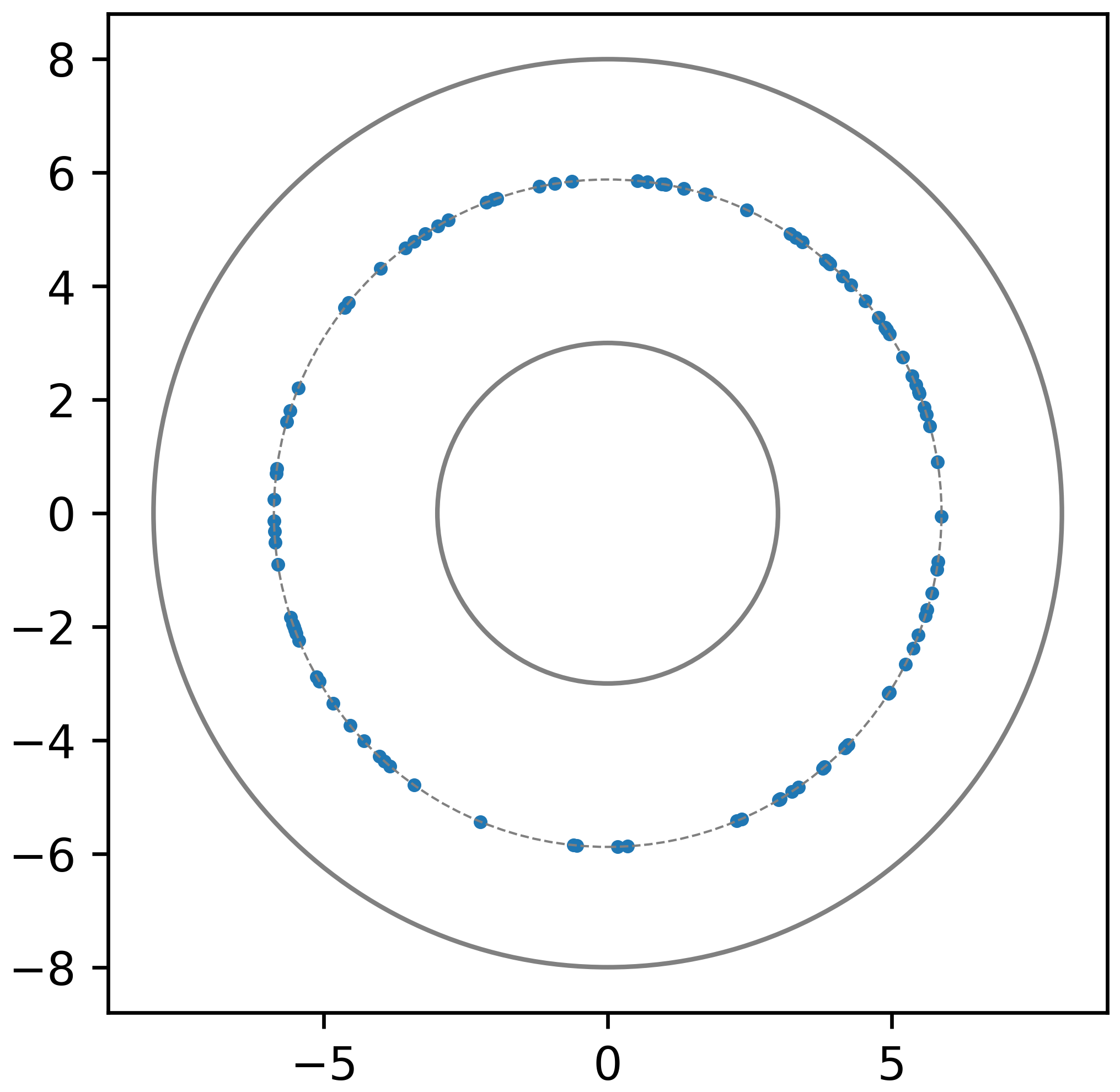}
      \caption{Injective flow approximation}
      \label{fig:donut-ae}
    \end{subfigure} \\
    \hfill
    \begin{subfigure}{.3\textwidth}
      \centering
      \includegraphics[width=.9\linewidth]{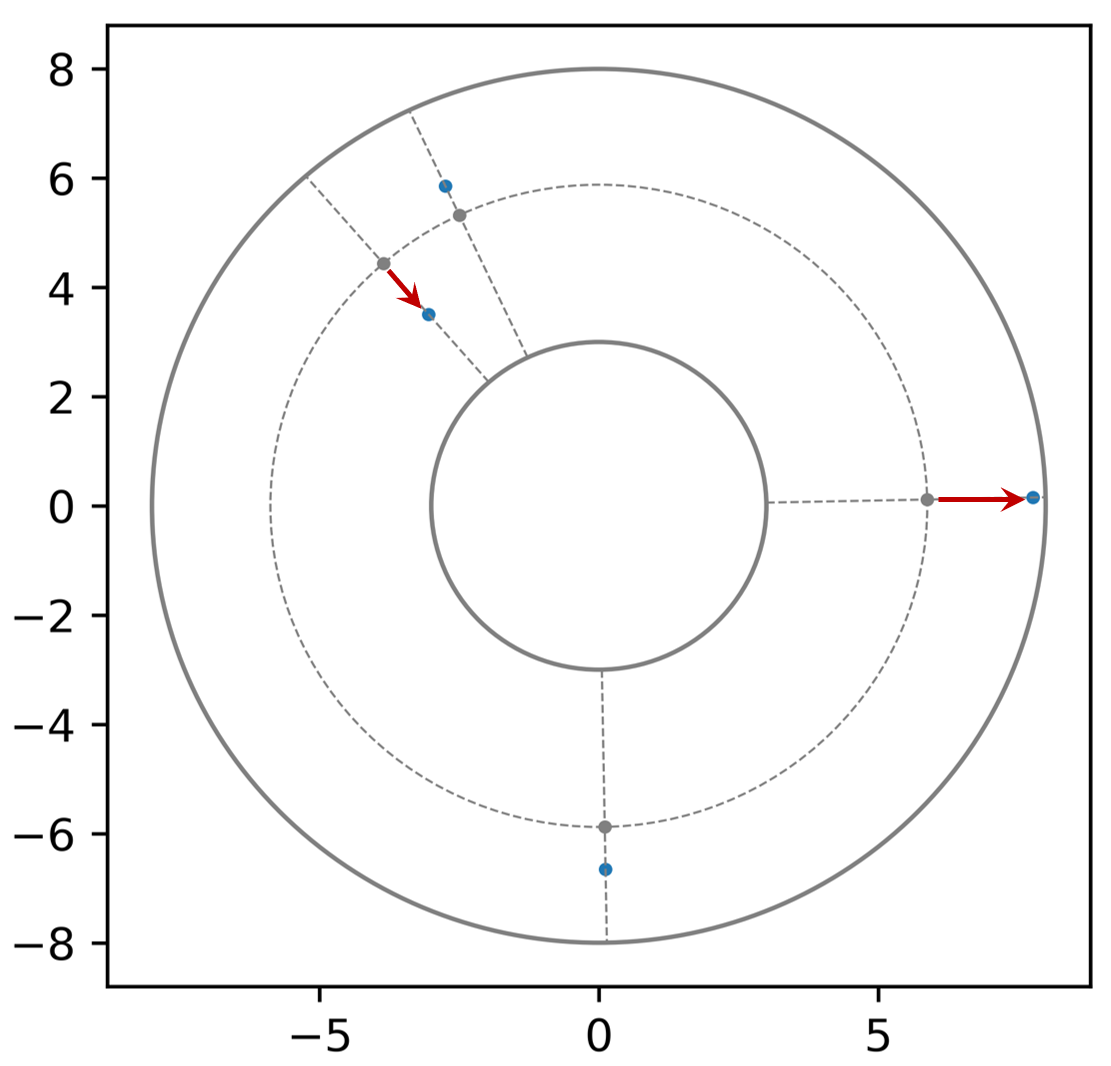}
      \caption{Split flow representation}
      \label{fig:donut-snm}
    \end{subfigure}%
    \begin{subfigure}{.3\textwidth}
      \centering
      \includegraphics[width=.91\linewidth]{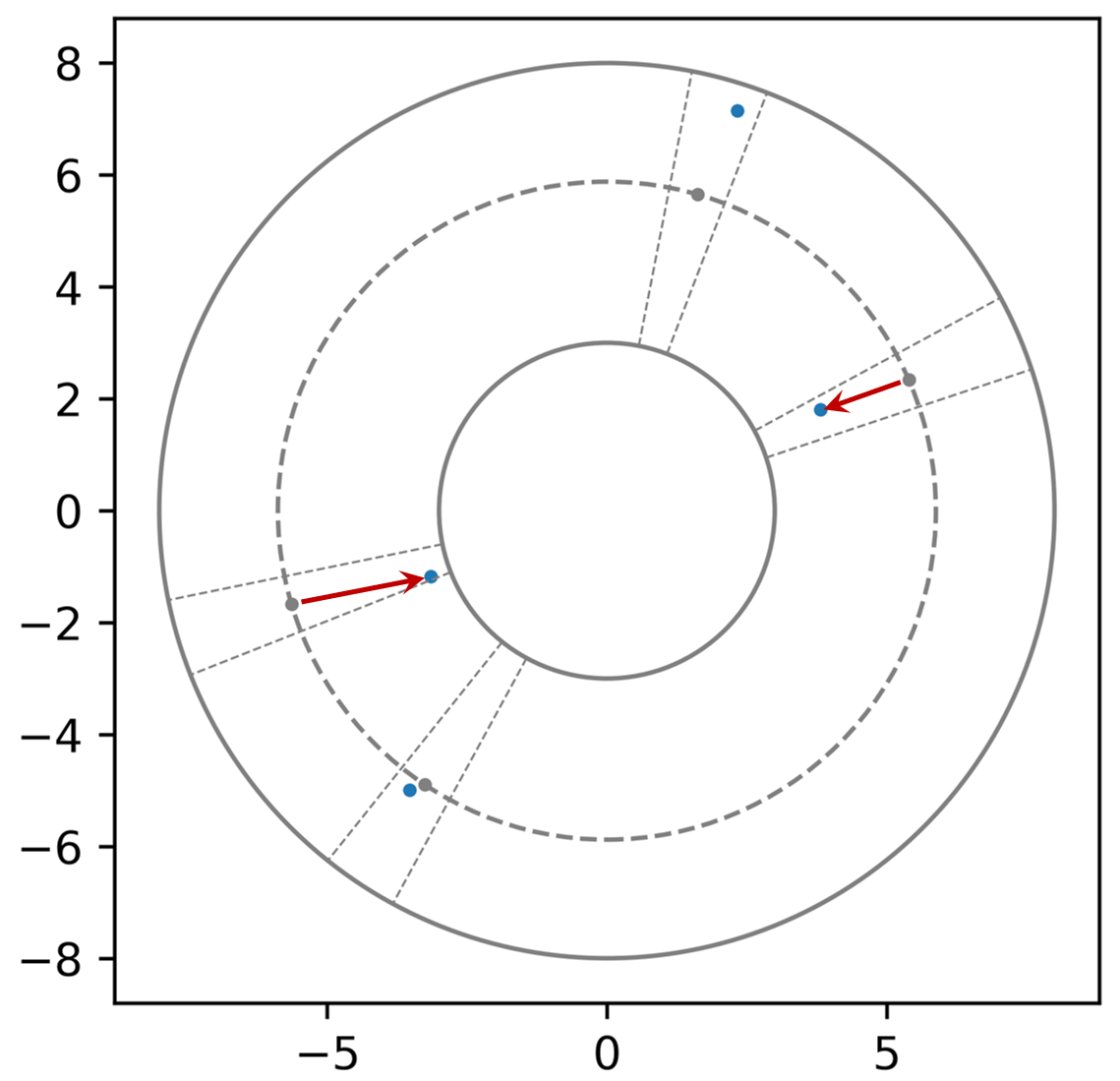}
      \caption{Stochastic flow representation}
      \label{fig:donut-sm}
    \end{subfigure}%
    \hspace{7.6mm}
    \caption{The uniform distribution on a donut (a) and various ways to represent or approximate it: (b)-(e).
    Description in the text.}
    \label{fig:donut}
\end{figure}

Let us illustrate the difference between these four types with a 2-dimensional Gaussian data distribution
\begin{equation}
    p^*(\X)=\normal\big(\0, \Sigmab\big)\qquad\text{with}\qquad\Sigmab=
    \begin{pmatrix}
        1 & 0\\
        0 & 1/4
        \end{pmatrix}
\end{equation}
\begin{enumerate}
    \item A bijective flow generates this distribution perfectly from a standard normal code distribution by an invertible deterministic transformation
    \begin{align}
        \text{decoder:}& \qquad &&[x_1,x_2] = g_\text{B}(z_1, z_2) = [z_1, z_2/2]\quad\text{with}\quad \z\sim\normal\big(\0,\eye_2\big) \nonumber \\
        \text{encoder:} & \qquad &&[z_1, z_2] = f_\text{B}(x_1, x_2) = [x_1, 2\, x_2] 
        \label{eq:bijective-gauss}
    \end{align}
    Moreover, any rotation of the code space (and any other distribution-preserving transformation of $\z$) will also result in a valid bijective flow.
    \item An injective flow can be realized as an autoencoder with 1-dimensional code space.
    Since the decoder is deterministic, it can only generate a 1-dimensional approximation of the data distribution.
    The squared reconstruction error is minimized when this approximation is aligned with the major axis of the data, i.e. the $x_1$ direction:
    \begin{align}
        \text{decoder:}& \qquad &&[x_1,x_2] = g_\text{I}(z) = [z, 0]\quad\text{with}\quad z\sim\normal\big(0,1\big) \nonumber \\
        \text{encoder:} & \qquad &&\qquad z = f_\text{I}(x_1, x_2) = x_1
        \label{eq:injective-gauss}
    \end{align}
    \item A stochastic flow defines a joint distribution of data and codes, such that the marginalization over codes results in the desired data distribution.
    A possible model with 1-dimensional codes is
    \begin{align}
        p(\X, Z) = \normal\big(\0, \Sigmab_\text{S}\big)\qquad\text{with}\qquad\Sigmab_\text{S}=
    \begin{pmatrix}
        1 & 0 & \rho\\
        0 & 1/4 & 0 \\
        \rho & 0 & 1
        \end{pmatrix}
    \end{align}
    where $\rho$ is the correlation between $X_1$ and $Z$.
    It is easy to see that this generates $p^*(\X)$ exactly.
    Encoder and decoder are no longer deterministic functions, but conditional probabilities:
    \begin{align}
        \text{decoder:}& \quad &&[x_1, x_2] \sim p_\text{S}(\X\given Z) = \normal\big([\rho\cdot z,\, 0], \Sigmab_\text{SD}\big)\quad\text{with}\quad\Sigmab_\text{SD}=
    \begin{pmatrix}
        1-\rho^2 & 0\\
        0 & 1/4
        \end{pmatrix} \nonumber \\
        \text{encoder:} & \quad &&\qquad z \sim p_\text{S}(Z\given\X) = \normal\big(\rho\cdot x_1,\, 1-\rho^2\big)
    \end{align}
    The marginal code distribution is again standard normal, $p(Z)=\normal(0, 1)$.
    The main characteristic of a stochastic flow is that the same $\x$ can be generated from (infinitely many) different $z$,
    or equivalently, that multiple codes can be assigned to any given $\x$.
    \item A split flow extends the autoencoder: it adds a second stochastic decoder stage in order to recover the missing dimension(s) arising from the code bottleneck and generate the full distribution $p^*(\X)$.
    The encoder and the first stage of the decoder from (\ref{eq:injective-gauss}) remain unchanged:
    \begin{align}
        \text{decoder 1:}& \qquad &&x_1 = g_\text{SP}(z) = z\quad\text{with}\quad z\sim\normal\big(0,1\big) \nonumber \\
        \text{decoder 2:}& \qquad &&x_2 \sim p_\text{SP}(X_2\given X_1) = \normal\big(0,\, 1/4\big) \nonumber \\
        \text{encoder:} & \qquad &&\qquad z = f_\text{SP}(x_1, x_2) = x_1
        \label{eq:split-gauss}
    \end{align}
    In this example, $x_2$ is independent of $x_1$, but this is a coincidence.
    A split flow differs from a stochastic flow in that it maintains a deterministic encoder and thus assigns exactly one code to each $\x$.
    Split flows can be derived from bijective flows by enforcing a split of the latter's code space into a core and a detail part.
    The code space in (\ref{eq:bijective-gauss}) happens to be ordered this way.
    Thus, the behavior of (\ref{eq:split-gauss}) can be replicated with our bijective flow by dropping the code variable $z_2$ after encoding and sampling a new value $z_2\sim\normal(0, 1)$ before decoding.
    Similarly, the behavior of the autoencoder in (\ref{eq:injective-gauss}) is obtained by setting $z_2=0$ before decoding.
    This nicely illustrates the different modes supported by split flows.
\end{enumerate}
As another toy example, consider a 2-dimensional uniform distribution over a donut with inner (interior) radius $R_0=3$ and outer (exterior) radius $R_1=8$, as depicted in \autoref{fig:donut-target}.
By construction, the true (uniform) density on the donut is the inverse of the area 
\begin{equation}
    p^*(\X) = \frac{1}{\pi(R^2_1 - R^2_0)}
\end{equation}
Once again, the four model types represent this distribution in fundamentally different ways. 
A bijective flow (\autoref{fig:donut-nf}) samples codes from a standard normal (gray dots) and transports them radially to their appropriate locations for the uniform donut distribution.
An injective flow (\autoref{fig:donut-ae}) implemented as a standard autoencoder with 1-dimensional bottleneck generates data on a 1-D subset by minimizing the squared reconstruction error. 
Due to the symmetry of the problem, the codes span the interval $[0,2\pi)$, and the reconstruction is a uniform distribution on a circle with radius $R_\MS\approx 5.88$.
Extending the autoencoder, a split flow (\autoref{fig:donut-snm}) first samples points uniformly on the circle, forming the data's ``core'' (gray), but then shifts these points radially by a second random draw from the ``detail'' distribution (blue).
Similarly, a stochastic flow (\autoref{fig:donut-sm}) first samples on the ``core'' circle (gray), but then shifts points in two dimensions within a wedge-shaped segment of the donut (blue).
The models (b), (d), and (e) all generate the target distribution exactly, whereas model (c) is an approximation.
We will derive the exact analytical change-of-variables formulas for these examples in \autoref{sec:donut-distribution}.

In this paper, we set out to present the various change-of-variables formulas in a systematic way.
Specifically, we will clarify the relationship between decoder architectures and induced CoV formulas, discuss efficient CoV computation, and point out some potentially new consequences for learning and inference.
We feel that interested researchers will profit from a unified and comprehensive presentation of the wide range of possibilities and important potential pitfalls that these formulas hide.
Readers already familiar with the classic CoV formulas for bijective mappings may want to focus on \autoref{sec:injective-flows} and \autoref{sec:split-flows}, which generalize these concepts to injective and split architectures. 

\section{Basic Concepts}\label{sec:basic-concepts}

{\bf Notation:} Upper-case letters $\X, \Z$ denote random vectors and lower-case letters $\x, \z$ their corresponding realizations (instances).
Upper-case letters are also used for matrices, but the distinction should always be clear from context.
The colon notation $\X_{i:j}$ selects the sub-vector of $\X$ with indices $i$ to $j$ (inclusive).
Vectors are understood as column vectors, and derivatives with respect to  a vector as row vectors, so that the Jacobian $\frac{\partial \x}{\partial \z}$ has dimension $\dim(\x) \times \dim(\z)$.
The operator $\diag(\x)$ constructs a diagonal matrix from the vector $\x$.
The symbol $\mathbbm{1}[\mathit{condition}] = 1 \text{ (if } \mathit{condition}=\text{true) and } 0 \text{ (if } \mathit{condition}=\text{false)}$ is the indicator function for the given condition.
The domain of a function or a probability measure is denoted by $\dom(.)$.
We write \smash{$\overset{!}{=}$} to indicate an equality that does not hold universally, but must be enforced by model design and/or training.
Similarly, \smash{$\overset{?}{=}$} denotes a hypothetical equality that turns out to be untrue in the present context.

We consider data that are represented as $D$-dimensional vectors $\x$.
The target data density $p^*(\X)$ shall be supported (i.e., non-zero) on some subset $\XS \subseteq \R^D$ of $D$-dimensional space.
With the exception of \autoref{sec:data-on-embedded-manifolds}, we assume that $p^*(\X)$ is non-degenerate -- i.e., the intrinsic dimension of $\X$ is equal to the embedding dimension $D$.
Likewise, codes $\z$ are $C$-dimensional vectors with $p(\Z)$ supported on $\ZS \subseteq \R^C$.
In addition, we consider finite code sets $Z\in\{1,...,K\}$ and corresponding discrete distributions $p(Z)$, which arise in classical methods such as histograms and $K$-means clustering.

A given decoder defines a generative data density $p(\X)$, and the modeling goal is to ensure that $p(\X) = p^*(\X)$ or at least $p(\X) \approx p^*(\X)$ with small error.
When the decoder is deterministic, $\x = g(\z)$, the induced generative distribution is the {\em pushforward} of $p(\Z)$ by $g(\z)$:
\begin{equation}\label{eq:pushforward}
    \x \sim p(\X) \qquad\Longleftrightarrow\qquad \x = g(\z) \ \text{ with }\ \z \sim p(\Z)
\end{equation}
A pushforward according to the above rule is commonly written with the $\#$-symbol as $p(\X) = g_\# p(\Z)$.
When the decoder is stochastic, $\x \sim p(\X\given\Z)$, the generative distribution is obtained by marginalization over $\Z$:
\begin{equation}
    p(\X) = \int_\ZS p(\X\given\Z\!=\!\z)\ p(\Z\!=\!\z)\ d\z
\end{equation}
To derive theory for these generative models and to train them in practice, it is useful to define complementary {\em encoders}, which invert the action of the decoders and transform data $\x$ into corresponding codes $\z$. 
The concrete meaning of ``invert'' depends on the specific type of decoder and is expressed by self-consistency requirements.
In the deterministic case, the encoder is a function $\z=f(\x)$ which fulfills
\begin{align}
    \forall \z \in \ZS:\quad \z \overset{!}{=} & \, f(g(\z))\quad\text{if }\dim(\ZS) \le \dim(\XS) \label{eq:deterministic-self-consistency} \\
    \forall \x \in \XS:\quad \x \overset{!}{=} & \, g(f(\x))\quad\text{if }\dim(\ZS) \ge \dim(\XS)
\end{align}
A stochastic encoder is represented by the conditional $\z \sim p(\Z\given\X)$.
Here, self-consistency means that the joint distributions of $\x$ and $\z$ induced by the encoder and decoder must be the same:
\begin{equation}\label{eq:bayes-self-consitency}
    p_E(\X,\Z) = p^*(\X)\ p(\Z\given\X) \overset{!}{=} p(\Z)\ p(\X\given\Z) = p_D(\X, \Z)
\end{equation}
This is the criterion optimized by the ELBO objective for variational autoencoder training. 
When encoder and decoder are separate models, the self-consistency formulas (\ref{eq:deterministic-self-consistency}) to (\ref{eq:bayes-self-consitency}) are often only approximately fulfilled, and the degree of deviation is an important indicator for model inaccuracy.

Note that we will express change-of-variables formulas in terms of probability densities $p(\X)$, whereas in practice they are mainly used in the equivalent, but numerically more robust, form of log-densities $\log p(\X)$.
All results can be trivially transferred to this convention.

\section{Bijective Flows}
\label{sec:bijective-flows}

Bijective flows are the canonical setting where change-of-variables formulas arise.
They are also known as {\em normalizing flows}, because the CoV formula ensures that the inferred density $p(\X)$ is properly normalized.
Both terms refer to models whose decoder is a deterministic bijective mapping from $\ZS$ to $\XS$.
Evidently, this requires $\dim(\XS)=\dim(\ZS)$ and leads to a lossless coding of $\x$ into $\z$ and back.
The concept of normalizing flows was introduced to machine learning in \autocite{rezende2015variational}, see  \autocite{papamakarios2021normalizing_flows, kobyzev2021normalizing_flows} for recent reviews.

\subsection{Finite Compositions: Bijective Decoder Functions}
\label{sec:finite-nfs}

Finite compositions represent bijective decoder functions $\x = g(\z)$ explicitly, e.g. by means of invertible neural networks that reduce complex bijective transformations to finite sequences of simpler bijective layers. 
For the model to be self-consistent, the encoder must realize the decoder's exact inverse $\z = f(\x) = g^{-1}(\x)$, which is guaranteed to exist thanks to bijectivity.
Many successful architectures avoid potential inconsistencies in the first place by implementing the encoder not as a separate function, but by simply running the decoder backwards. 

The most popular choice for the code distribution is the standard normal,  $p(\Z) = \normal(0, \eye_C)$, but other code distributions have been successfully used as well, for example Student's t \autocite{jaini2020tails,alexanderson2020robust}, tensor trains \autocite{khoo2023tensorizing}, or Gaussian mixtures (see below and \autoref{sec:mixture-models}).
The crucial requirement is that $p(\Z)$ must be known and tractable.
The generative data distribution is then given by the pushforward (\ref{eq:pushforward}), resulting in the well-known bijective change-of-variables formula (a straightforward consequence of the general change-of-variables formula of integration)
\begin{flalign}\label{eq:bijective-cov}
    \CoV{bijective CoV}&&
    p(\X\!=\!\x) = p(\Z\!=\!f(\x))\ \big|\det\big(\J_f(\x)\big)\big| = p(\Z\!=\!f(\x))\ \big|\det\big(\J_g(f(\x))\big)\big|^{-1}&&
\end{flalign}
where $\J_f$ and $\J_g$ are the Jacobians of the encoder and decoder functions respectively
\begin{equation}
    \J_f(\x)\, = \, \left.\frac{\partial f(\x)}{\partial \x} \right|_{\x}  \qquad\qquad \J_g(\z)\, = \, \left.\frac{\partial g(\z)}{\partial \z} \right|_{\z}
\end{equation}
and their determinants are always non-zero because $f(\x)$ and $g(\z)$ are invertible by construction.
Important realizations for this model class include {\em triangular maps}, for example autoregressive flows \autocite{huang2018neural,papamakarios2017masked}, Knothe-Rosenblatt rearrangements \autocite{marzouk2016sampling}, and structural causal models \autocite{xia2021causal,khemakhem2021causal},
as well as {\em invertible ResNets} \autocite{behrmann2019invertible,chen2019residual}, and {\em coupling flows} \autocite{dinh2017density,kingma2018glow}.
These approaches differ in how they ensure that the Jacobian determinants in (\ref{eq:bijective-cov}) are efficiently computable.
For instance, triangular maps are constructed such that the Jacobian is a triangular matrix and its determinant is therefore just the product of the diagonal elements.
The other approaches exploit the fact that the Jacobian determinant of a multi-layered architecture $\x = g(\z) = (g_L \circ ... \circ g_1)(\z)$, with intermediate variables defined by $\z^{(l)}=g_l\big(\z^{(l-1)}\big)$ and $\z^{(0)} = \z$, $\x = \z^{(L)}$, is the product of the determinants of the individual layers
\begin{equation}
    \det\big(\J_g(\z)\big) = \prod_{l=1}^L \det\big(\J_{g_l}\big(\z^{(l-1)}\big)\big)
\end{equation}
The calculation of the individual determinants is made efficient by keeping them constant (linear layers), expanding them into a truncated series (residual layers), or enforcing triangular Jacobians (coupling and actnorm layers).
More details on the calculation of Jacobian determinants are provided in \autoref{sec:efficient-jacobians}.
The complications of efficient determinant computation are avoided by {\em incompressible flows}, which are designed such that $\det(\J_f)=\det(\J_g)=1$ for all $\x$ and $\z$, and thus we obtain the 
\begin{flalign}\label{eq:incompressible-cov}
    \CoV{incompressible bijective CoV} 
    && \hspace{-3cm} p(\X=\x)=p(\Z=f(\x))  &&
\end{flalign}
Incompressible flows can be realized by restricting $g(\z)$ to an isometry (e.g., $\x = \Q \cdot \z + \mub$ with orthogonal matrix $\Q$ as in PCA, inverse Fourier, or wavelet transforms), or by using special architectures that guarantee unit determinants, such as NICE \autocite{dinh2014nice} or GIN \autocite{sorrenson2019disentanglement}.
However, incompressible flows are much less expressive than unconstrained ones.
For example, they are unable to change the number of modes between $p(\Z)$ and $p(\X)$. 

On the other hand, model expressiveness can be increased by more complex code distributions.
Two choices are popular: Gaussian mixture models (GMMs) and vector quantization (VQ).
In {\em GMM flows} \autocite{sorrenson2019disentanglement,izmailov2020semi,ardizzone2020training,hagemann2021stabilizing,willetts2021don}, the code distribution is defined as
\begin{equation}
    p(\Z\!=\!\z) = \sum_{k=1}^K \,p(k)\cdot \normal(\z\given\mub_k,\Sigmab_k)
\end{equation}
The mixture weights $p(k)$ as well as the means $\mu_k$ and covariances $\Sigmab_k$ of the mixture components must be learned along with the flow $g(\z)$.
To reduce the number of learnable parameters, the covariances are usually restricted to be diagonal.
The resulting change-of-variables formula is
\begin{flalign}\label{eq:gmm-flow-cov}
    \CoV{GMM flow CoV} && \hspace{-2mm}
    p(\X\!=\!\x) = \big| \det\big(\J_f(\x)\big)\big| \cdot \sum_{k=1}^K\, p(k)\cdot \normal\big(f(\x)\given \mub_k,\Sigmab_k\big)  &&
\end{flalign}
GMM flows are especially useful when the data instances have natural labels, such as MNIST digits or ImageNet categories.
This allows for supervised learning of the latent structure: 
Instances of class $k$ only influence the corresponding mixture component, while all classes share the same flow $f(\x)$.
However, the loss may diverge when the covariances of the mixture components are also learned.
This can be avoided by using an incompressible flow \autocite{sorrenson2019disentanglement}.
Attempts at unsupervised learning of GMM flows are inconclusive \autocite[e.g., Appendix F in][]{willetts2021don}.

Instead of soft assignments of data points to mixture components, differentiable tesselations \autocite{chen2022semi} and VQ flows \autocite{sidheekh2022vq} use hard cluster assignments similar to $K$-means%
\footnote{\textcite{sidheekh2022vq} also propose a variant with overlapping clusters, which we do not delve into here. 
Their paper does not specify which variant is used in the experiments.}.
To this end, VQ flows train an additional cluster assignment function $k=h(\x)$ that returns the (unique) cluster label of instance $\x$.
Then, separate normalizing flows $f_k(\x)$ are learned for each cluster.
We will discuss this method in more detail in \autoref{sec:data-on-embedded-manifolds} and only report the resulting change-of-variables formula here:
\begin{flalign}\label{eq:vq-flow-cov}
    \CoV{VQ flow CoV} && \hspace{-2mm}
    p(\X\!=\!\x) = \sum_{k=1}^K\,\mathbbm{1}[h(\x)\!=\!k]\cdot p(k)\cdot p\big(\Z\!=\!f_k(\x)\big) \cdot \big| \det\big(\J_{f_k}(\x))\big)\big|    &&
\end{flalign}
This method is especially useful when the data distribution $p^*(\X)$ consists of multiple disconnected components.
Standard normalizing flows struggle to represent such distributions because they lead to exploding or vanishing Jacobian determinants.
However, the main use case of VQ flows is the modeling of embedded manifolds, which we treat in \autoref{sec:data-on-embedded-manifolds}.

\subsection{Infinitesimal Compositions: Deterministic Diffusion Processes}\label{sec:neural-ode}

In this setting, encoder and decoder are not defined by a bijective neural network, but by a continuous diffusion process that implicitly transforms between the target distribution $p(\X)$ and the latent distribution $p(\Z)$ \autocite{chen2018neural,grathwohl2019ffjord}.
Note that we reverse the time direction used in these papers in order to maintain consistency with the Markov chain and stochastic differential equation models in \autoref{sec:stochastic-models}.
The encoder of an infinitesimal normalizing flow is defined by an ordinary differential equation (ODE)
\begin{equation}
    \frac{d}{d t}\z(t) = F\big(t, \z(t)\big)
\end{equation}
whose RHS is a learnable vector field determining the infinitesimal flow at time $t$.
With initial condition $\z(0) = \x \sim p^*(\X)$, the encoder transformation is implicitly defined by integrating the ODE from $t=0$ to $t=T$:
\begin{equation}
    \z = f(\x): = \z(0) + \int_0^T F\big(t, \z(t)\big)\, dt
\end{equation}
The corresponding decoder $\x = g(\x)$ is obtained by integrating the ODE backwards in time with initial condition $\z(T) = \z \sim p(\Z)$;
\begin{equation}
    \x = g(\z) := \z(T) + \int_T^0 F\big(t, \z(t)\big)\, dt 
\end{equation}
The generative distribution is thus the pushforward $p(\X) = g_\#p(\Z)$, and the vector fields $F\big(t, \z(t)\big)$ are trained such that $p(\X) \approx p^*(\X)$.
The encoder's infinitesimal change in log-density is given by the trace of the vector field's Jacobian, the so-called {\em instantaneous change-of-variables formula} \autocite{chen2018neural,grathwohl2019ffjord}, which follows naturally from the continuity equation of fluid mechanics: 
\begin{equation}
    \frac{d}{d t}\log p\big(\Z_t=\z(t)\big) = \tr\big(J_F(t, \z(t))\big)
    \quad\text{with}\quad
    J_F(t, \z(t)) = \left.\frac{\partial F(t, \z)}{\partial \z} \right|_{t,\z(t)}
\end{equation}
Hence, the total change-of-variables of a continuous flow can be obtained by integrating this expression forwards in time, much like the flow itself:
\begin{flalign}
    \CoV{continuous CoV} \label{eq:continuous-cov} && \hspace{-1cm}
    p(\X=\x) = p\big(\Z=f(\x)\big)\cdot  \exp\left(\int_0^T\! \tr\big(J_G(t, \z(t))\big)\, dt\right) &&
\end{flalign}
Methods for efficient computation of $\tr\big(J_F(t, \z(t))\big)$ are discussed in \autoref{sec:efficient-jacobians}.

\todo{discuss Riemannian flows \autocite{mathieu2020riemannian}, neural ODEs on manifolds \autocite{falorsi2020neural}?}

\section{Stochastic Flows}\label{sec:stochastic-models}

Stochastic flows are characterized by the property that a given $\x$ is not assigned to a unique code $f(\x)$, but to a set of codes drawn from a conditional encoder distribution $\z \sim p(\Z\given\X\!=\!\x)$.
This leads to the important consequence that the joint encoder distribution $p_E(\X, \Z) = p(\X)\,p(\Z\given\X)$ is a non-degenerate density, whereas a bijective encoder has a degenerate joint density $p_E(\X, \Z) = p(\X)\,\delta\big(\z - f(\x)\big)$.
In order to achieve self-consistency according to equation (\ref{eq:bayes-self-consitency}), the decoder must now be stochastic as well, and generate the same joint distribution
\begin{equation}
    p_D(\X, \Z) = p(\Z)\,p(\X\given\Z)\overset{!}{=}p_E(\X, \Z) = p(\X)\,p(\Z\given\X).
\end{equation}
A straightforward possibility to define the generative distribution $p(\X)$ of a stochastic flow is to marginalize the joint decoder distribution $p_D(\X, \Z)$ over $\Z$
\begin{flalign}\label{eq:decoder-marginalization-cov}
    \CoV{decoder marginalization CoV} && \hspace{-20mm}
    p(\X\!=\!\x) = \E_{\z\sim p(\Z)}\big[p(\X\!=\!\x\given\Z\!=\!\z)\big]  &&
\end{flalign}
However, this expression -- although simple at first glance -- is often intractable in practice.
Below, we first discuss an important tractable variant, the discrete mixture models, and then consider alternative expansions of $p(\X)$ that are numerically simpler. 

\subsection{Mixture Models}\label{sec:mixture-models}

When the codes are discrete, $Z\in\{1,...,C\}$, the marginalization in equation (\ref{eq:decoder-marginalization-cov}) reduces to a simple sum.
Then, $p(\X)$ can be expressed as a Gaussian mixture model (GMM), such that the codes index Gaussian conditionals $p(\X\!=\!\x\given Z\!=\!k)=\normal(\x\given \mub_k, \Sigmab_k)$ with the code distribution $p(Z\!=\!k)$ corresponding to the mixture weights. 
Since Gaussians have infinite support, each mixture component contributes to every data point, resulting in the change-of-variables formula
\begin{flalign}\label{eq:gmm-cov}
    \CoV{GMM CoV} && \hspace{6mm}
    p(\X\!=\!\x) = \sum_{k=1}^C p(Z\!=\!k)\cdot \normal(\x\given \mub_k, \Sigmab_k)  &&
\end{flalign}
The encoder corresponds to the posterior $p(Z\!=\!k\given\X\!=\!\x)$, where the probability of each code $k$ expresses the degree of membership of $\x$ in the corresponding component $\normal(\x\given \mub_k, \Sigmab_k)$.
Self-consistency is therefore automatically ensured when the encoder is implicitly defined via Bayes rule 
\begin{equation}
p(Z\!=\!k\given\X\!=\!\x)=p(Z\!=\!k)\cdot \normal(\x\given \mub_k, \Sigmab_k)\, /\, p(\X\!=\!\x)
\end{equation}
It may be surprising to see the well-known GMM equation (\ref{eq:gmm-cov}) interpreted as a change-of-variables formula, but the relationship between $\X$ and $Z$ conforms precisely to our definition of a stochastic encoder/decoder architecture.
We will see later that this viewpoint also sheds interesting new light on other classical models.

The ideas of GMMs and NFs can be combined in {\em GMM flows}, that is, NFs whose code distribution is a GMM instead of a standard normal.
Even though we already introduced these models in \autoref{sec:finite-nfs}, we take a slightly different perspective here:
In line with the present discussion, we use the GMM's mixture component labels to define a finite set of codes $Z\in\{1,...,C\}$.
Thus, we must rename the latent variable of the flow (which acted as a code in \autoref{sec:finite-nfs}) into $\S_x$.
Expressed in this way, GMM flows turn into stochastic flows over the joint distribution of data and codes.
Re-writing them in terms of a stochastic decoder, conditioned on the discrete codes, gives
\begin{equation}\label{eq:latent-mixture-pushforward}
    \x \sim p(\X\given Z\!=\!k) \quad\,\Longleftrightarrow\quad\, \x = g(\s_x) \ \text{ with }\ \s_x\sim p(\S_x\given Z\!=\!k)=\normal(\S_x\given \mub_k,\Sigmab_k)
\end{equation}
Of course, this re-interpretation does not change the marginal data distribution $p(\X)$, which is still given by the change-of-variables formula (\ref{eq:gmm-flow-cov}).
However, it shows that the decoder of a GMM flow can be interpreted as a stochastic flow with joint distribution $p_D(\X\!=\!\x, Z\!=\!k) = p(Z\!=\!k)\cdot\normal(\S_x\!=\!g^{-1}(\x)\given \mub_k,\Sigmab_k)$.


\subsection{Bayesian Models and Markov Chains}\label{sec:bayesian-models}

Bayesian models\footnote{Not to be confused with Bayesian neural networks, which are characterized by having a weight distribution instead of fixed weights.} learn encoder/decoder pairs that explicitly implement the complementary conditionals $p(\Z\given\X)$ and $p(\X\given\Z)$.
In contrast to bijective transformations, the code dimension of a Bayesian model can be arbitrary, because valid conditionals $p(\Z\given\X)$ and $p(\X\given\Z)$ can be formulated for any choice of the space $\ZS$.
Indeed, it is permitted for $\X$ and $\Z$ to be statistically independent, a degeneracy that occasionally occurs in the training of certain variants of this model type \autocite{chen2017variational,zhao2019infovae}.

If the self-consistency requirement (\ref{eq:bayes-self-consitency}) is fulfilled, we can avoid the intractable marginalization (\ref{eq:decoder-marginalization-cov}) and instead write $p(\X)$ by rearranging Bayes' rule:
\begin{flalign}\label{eq:bayes-cov}
    \CoV{Bayesian CoV} && \hspace{-6mm}
    p(\X\!=\!\x) = p(\Z\!=\!\z)\,\frac{p(\X\!=\!\x\given\Z\!=\!\z)}{p(\Z\!=\!\z\given\X\!=\!\x)}  &&
\end{flalign}
This change-of-variables formula combines encoder and decoder instead of marginalizing over the decoder alone.
It reduces to the bijective CoV formula (\ref{eq:bijective-cov}) when the complementary conditionals are taken to their deterministic limits $p(\X\!=\!\x\given\Z\!=\!\z) \rightarrow \delta\big(\x-g(\z)\big)$ and $p(\Z\!=\!\z\given\X\!=\!\x) \rightarrow \delta\big(\z -g^{-1}(\x)\big)$, as was proven in the supplements of \autocite{wu2020stochastic} and \autocite{nielsen2020survae}.

Importantly, the Bayesian CoV formula highlights a crucial requirement for valid stochastic codes: 
It must hold for {\em all} $\z$ in the domain $\dom\!\big(p(\Z\given\X\!=\!\x)\big)$ of the encoder conditional.
In other words, since the LHS of (\ref{eq:bayes-cov}) is a constant for fixed $\x$, the RHS must return the {\em same value for any} $\z\sim p(\Z\given\X\!=\!\x)$.
This is another way of expressing the self-consistency of stochastic encoder/decoder pairs and gives raise to the famous ELBO (``evidence lower bound'') objective.
The ELBO follows from applying Jensen's inequality to the logarithm of (\ref{eq:decoder-marginalization-cov})
\begin{align}
    \log p(\X) &= \log \int p(\Z)\cdot p(\X\given\Z)\,d\Z = \log \int p(\Z\given\X)\cdot \frac{p(\Z)\cdot p(\X\given\Z)}{p(\Z\given\X)}\,d\Z \nonumber \\
    &\ge \int p(\Z\given\X)\cdot \log\frac{p(\Z)\cdot p(\X\given\Z)}{p(\Z\given\X)}\,d\Z =: \text{ELBO}
\end{align}
It is easy to show that the difference between LHS and RHS of this inequality is the KL divergence between the encoder $p(\Z\given\X)$ and the decoder posterior $p_D(\Z\given\X)=p(\Z)\cdot p(\X\given\Z)\, /\, p(\X)$
\begin{align}
    \log p(\X) - ELBO = \kl\big[p(\Z\given\X) \,||\, p_D(\Z\given\X)\big]
\end{align}
This difference should be zero in a self-consistent stochastic flow. 
It can be tractably estimated by the second order Taylor expansion of the KL divergence, which gives
\begin{equation}
    \hspace{-2mm} \kl\big[p(\Z\given\X\!=\!\x) \,||\,\, p_D(\Z\given\X\!=\!\x)\big]\approx \frac{1}{2\,p(\X\!=\!\x)^2}\,\Var_{\z\sim p(\Z\given\X\!=\!\x)}\!\!\left[p(\Z\!=\!\z)\frac{p(\X\!=\!\x\given\Z\!=\!\z)}{p(\Z\!=\!\z\given\X\!=\!\x)}\right]
\end{equation}
In other words, the variance of the RHS of the Bayesian CoV formula (\ref{eq:bayes-cov}) over $\z\sim p(\Z\given\X\!=\!\x)$ is a powerful diagnostic for self-consistency:
When the variance is too big for some or all $\x$, the model has not converged to a self-consistent state and needs to be improved.
This criterion complements the goodness-of-fit requirement $\log p(\X)\approx \log p^*(\X)$ and has, for example, been demonstrated in \autocite{radev2023likelihood_posterior}.

Instead of a single code space, we can specify a sequence of code spaces $\ZS_1,...\ZS_T$.
Then, we can define multi-layer decoders and encoders as {\em Markov chains}:
\begin{align}
    p_D(\X, \Z_1,...,\Z_T) =&\,p(\X \given \Z_1)\cdot p(\Z_1 \given \Z_2)\cdot ... \cdot p(\Z_{T-1}\given \Z_T)\cdot p(\Z_T)
    \label{eq:markov-chain-decoder} \\
    p_E(\X, \Z_1,...,\Z_T) =&\,p(\Z_T\given \Z_{T-1})\cdot ... \cdot p(\Z_2\given \Z_1) \cdot p(\Z_1\given \X)\cdot p(\X)
    \label{eq:markov-chain-encoder}
\end{align}
The existence of the encoder chain $p_E(\X, \Z_1,...,\Z_T)$ for a given decoder chain $p_D(\X, \Z_1,...,\Z_T)$ is guaranteed under mild assumptions \autocite{hagemann2022stochastic}.
To generalize the Bayes CoV formula (\ref{eq:bayes-cov}), we need one ratio term for each complementary pair of conditionals.
For notational brevity, we identify $\Z_0\equiv\X$:
\begin{flalign}\label{eq:markov-chain-cov}
    \CoV{Markov chain CoV} && \hspace{-6mm}
    p(\X\!=\x) = p(\Z_T\!=\!\z_T)\, 
    \prod_{t=1}^{T}\, \frac{p(\Z_{t-1}\!=\!\z_{t-1}\given\Z_t\!=\!\z_t)}{p(\Z_t\!=\!\z_t\given\Z_{t-1}\!=\!\z_{t-1})}  &&
\end{flalign}
Self-consistency implies that the RHS of this formula gives the same value for {\em every} execution path $\z_1, ..., \z_T$ of the encoder chain when the initial state is fixed to $\X\!=\x$.

A one-step Bayesian encoder/decoder pair is recovered by marginalizing over $\Z_1,...,\Z_{T-1}$ and equating $\Z:=\Z_T$.
Conversely, letting $T\rightarrow \infty$ results in a continuous-time model expressed as a stochastic differential equation, see \autoref{sec:stochastic-differential-equations}.
An in-depth treatment of Markov chain encoders/decoders can be found in \autocite{hagemann2022stochastic} under the name of {\em stochastic normalizing flows}.
Interesting special cases are considered in \autocite{wu2020stochastic}, who propose normalizing flows with alternating deterministic (coupling) and stochastic layers and demonstrate the superior expressive power of this architecture.
A key contribution of their work is the derivation of simple analytical expressions for the change-of-variable contribution of individual stochastic layers.

An interesting example for a Markov chain decoder is the {\em StyleGAN} \autocite{karras2019style}.
Its chain consists of 19 stages.
A latent code vector $\z \sim p(\Z)$ is first transformed into a style vector $\w = g_w(\z)$ of the same dimension via a deterministic push-forward.
This is followed by 18 convolutional layers $\z_{t-1} = g_t(\z_t, \w, \s_t)$ that gradually upscale the image from resolution $4\times 4$ to $1024\times 1024$ (two layers per resolution). 
The initial image $\z_{18}$ is fixed after training, and the final image $\x:=\z_0$ is the generated output.
The $\s_t\sim p(\S_t)$ are single channel noise images of the appropriate resolutions, which ensure diversity of the generated data.
Thus, each convolutional layer represents a conditional $p_t(\Z_{t-1}\given \Z_t,\W)$ with outsourced noise $\S_t$. 
Due to the deterministic relationship between $\Z$ and $\W$, these conditionals can be equivalently expressed as $p_t(\Z_{t-1}\given \Z_t,\Z)$, and the entire decoder chain becomes
\begin{equation}
    p(\X, \Z_1,...,\Z_{17}, \Z) = p(\Z)\cdot p_{18}(\Z_{17}\given \Z)\cdot \prod_{t=1}^{17} p_t(\Z_{t-1}\given \Z_t, \Z)
\end{equation}
Unfortunately, it is hard to transform this expression into a change-of-variables formula, because marginalization over the $\Z_t$ is intractable, and the complementary conditionals $p_t(\Z_t\given \Z_{t-1},\Z)$ for the encoder chain are unknown, so that (\ref{eq:markov-chain-cov}) is not applicable.

\subsubsection{Variational Autoencoders}

A canonical example for the Bayesian encoder/decoder architecture is the {\em variational autoencoder} \autocite[VAE,][]{kingma2014auto}, which typically incorporates a bottleneck $\dim(\Z) < \dim(\X)$.
Its marginal code distribution is standard normal, and the encoder and decoder conditionals are typically defined as diagonal Gaussian distributions whose parameters $\mub_E(\x), \sigmab_E(\x)$ resp. $\mub_D(\z), \sigmab_D(\z)$ are deterministic functions, i.e. the model implements a mean field approximation.
The Bayesian CoV formula (\ref{eq:bayes-cov}) then specializes into
\begin{flalign}\label{eq:vae-cov}
    \CoV{VAE Bayesian CoV} && 
    p(\X\!=\x) = \normal(\z\given \mathbf{0},\eye)\,\,\frac{\normal\big(\x\given\mub_D(\z), \diag(\sigmab^2_D(\z))\big)}{\normal\big(\z\given\mub_E(\x), \diag(\sigmab^2_E(\x))\big)}  &&
\end{flalign}
where once again the RHS must be constant for {\em all} $\z$ and fixed $\x$.
On the other hand, when $\sigmab^2_D$ and $\sigmab^2_E$ are small, code and data variability can be effectively neglected, which turns a VAE into an approximation of a deterministic autoencoder, see \autoref{sec:autoencoders}.

\subsubsection{Paired Conditional Normalizing Flows}

The VAE's restriction to diagonal Gaussians can be avoided by more expressive architectures, in particular {\em conditional normalizing flows} \autocite{trippe2018conditional,lu2020structured,winkler2019learning,ardizzone2020conditional}.
Conditional NFs realize a conditional pushforward similar to (\ref{eq:latent-mixture-pushforward}), but instead of conditioning the latent distribution, they use $\z$ to parameterize the transformation $g(.)$:
\begin{equation}\label{eq:conditional-pushforward}
    \x \sim p(\X\given\Z\!=\!\z) \qquad\Longleftrightarrow\qquad \x = g(\s_x; \z) \ \text{ with }\ \s_x\sim p(\S_x)
\end{equation}
Again, $\S_x$ denotes the flow's latent variable with $\dim(\S_x) = \dim(\X)$ and known distribution $p(\S_x)$, usually standard normal.
In contrast to the GMM representation of the code distribution in equation (\ref{eq:latent-mixture-pushforward}), $\Z$ is no longer restricted to be discrete.
The function $g(\s_x;\z)$ must be invertible with respect to its first argument, but not the second.
This existence of this conditional pushforward is guaranteed by the strong functional representation lemma \autocite{li2018strong}.
It generalizes the VAE's reparametrization trick and has also been termed ``noise outsourcing'' \autocite{bloem2020probabilistic}, because the uncertainty in $\X$ that remains after fixing $\z$ is ``delegated'' to the auxiliary variable $\S_x$ via the inverse $g^{-1}(\x;\z)$.
The decoder distribution can now be expressed by a conditional change-of-variables formula
\begin{flalign}\label{eq:condnf-cov}
    \CoV{conditional bijective CoV} && 
    p(\X\!=\!\x\given\Z\!=\!\z)\, =\, p\big(\S_x\!=\! g^{-1}(\x; \z)\big)\cdot \big| \det\big(\J_g(\x; \z))\big)\big|^{-1} &&
\end{flalign}
with Jacobian
\begin{equation}
    \J_g(\x,\z)\, = \, \left.\frac{\partial g(\s_x; \z)}{\partial \s_x} \right|_{\s_x=g^{-1}(\x;\z)}
\end{equation}
The marginal $p(\X)$ of this encoder is easy to compute when $p(Z)$ is discrete, for instance, when $Z\in\{1,...,C\}$ represents a finite number of class labels, as in latent mixture NFs.
However, when $\Z$ is continuous, defining $p(\X)$ in terms of the marginalization (\ref{eq:decoder-marginalization-cov}) is usually intractable. 

Instead, one defines a complementary encoder $p(\Z\given\X\!=\!\x)$ by another conditional NF. 
It realizes the pushforward $\z=f(\s_z; \x)$, where $\s_z\sim p(\S_z)$ denotes the outsourced noise variable corresponding to $\Z$ with $\dim(\S_z) = \dim(\Z)$.
Note that the functions $g$ and $f$ are no longer inverses of each other, contrary to the unconditional case.
Using the two CoV formulas for encoder and decoder, equation (\ref{eq:bayes-cov}) specializes into
\begin{equation}\label{eq:condnf-bayes-cov}
   \hspace{-5mm} \CoV{conditional NF Baysian CoV} \quad
    p(\X\!=\!\x) = p(\Z\!=\!\z)\,\frac{p\big(\S_x\!=\! g^{-1}(\x; \z)\big)\cdot \big| \det\big(\J_g(\x; \z))\big)\big|^{-1}}{p\big(\S_z\!=\! f^{-1}(\z; \x)\big)\cdot \big| \det\big(\J_f(\z; \x))\big)\big|^{-1}} 
\end{equation}
We are not aware of existing work where an encoder/decoder pair of conditional normalizing flows is used for unsupervised learning of codes, but the method shows great promise as a {\em supervised} learning approach in the context of {\em simulation-based inference} \autocite[SBI,][]{cranmer2020frontier}:
One can interpret the function $g(.)$ in (\ref{eq:conditional-pushforward}) as a simulation program with outcome $\x$ (e.g. a system of ordinary differential equations, ODEs).
The variable $\z$ here assumes the role of adjustable simulation parameters with prior $p(\Z)$ (e.g. the ODEs' coefficients and initial conditions), and $\s_x$ is a random vector with distribution $p(\S_x)$ accounting for observation noise.
In principle, the simulation could be directly used as a probabilistic decoder, but the Jacobian in (\ref{eq:condnf-cov}) is typically intractable.%
\footnote{This may change in the future with improved understanding of rectangular Jacobians and the advancement of {\em differentiable programming} in languages like JAX \autocite{jax2018github}, Julia/LLVM \autocite{moses2020rewriting}, Taichi \autocite{hu2019difftaichi}, and SWIFT \autocite{wei2021differentiable}, which aim at providing automated differentiation for all programming constructs including dynamic control flow. An early usage example is \autocite{wagner2022inverse}.}
Instead, one can use the simulation to generate a training set of parameter-outcome pairs $\{(\z_i,\x_i)\}_{i=1}^N$ for supervised learning of a conditional NF encoder/decoder pair.
The resulting decoder $p(\X\given\Z)$ serves as a probabilistic surrogate for the original simulation, and the encoder $p(\Z\given\X)$ represents the posterior distribution of parameters, given observations.
This technique was explored in depth by \autocite{radev2023likelihood_posterior}, who show that conditional NF pairs can achieve high degrees of self-consistency.


\subsection{Stochastic Differential Equations}
\label{sec:stochastic-differential-equations}

Stochastic differential equations (SDEs) are the continuous counterpart of Markov chains, in the same way as diffusion by ordinary differential equations (see \autoref{sec:neural-ode}) is the continuous version of normalizing flows.
This approach starts by defining an encoder that gradually adds Gaussian noise to the data via a Wiener process $d\w$ \autocite{ho2020denoising,song2021score}
\begin{equation}
    d\z = \Phi_\text{drift}(t, \z(t))\, dt + \Phi_\text{diff}(t)\, d\w
\end{equation}
where $\Phi_\text{drift}(t, \z(t))$ is a deterministic drift vector field (analogous to the ODE formulation), and $\Phi_\text{diff}(t)$ is the diffusion coefficient. 
The diffusion is initialized at $\z(t\!=\!0)=\x\sim p^*(\X)$ and runs forward to $t=T$.
The decoder is the corresponding reverse-time SDE, given by 
\begin{equation}
    d\z = \big[\Phi_\text{drift}(t, \z(t)) - \Phi_\text{diff}(t)^2\,\nabla_{\z} \log p_t(\z(t))\big]\, dt + \Phi_\text{diff}(t)\, d\tilde\w
\end{equation}
with negative time step $dt$ and reverse Wiener process $d\tilde\w$.
The term $\nabla_{\z} \log p_t(\z(t))$ is the score of the marginal distribution at time $t$.
To make this problem tractable, drift term and diffusion coefficient are often chosen such that the encoder SDE is analytically solvable.
As a popular example, we describe the {\em denoising diffusion probabilistic models} \autocite[DDPMs,][]{sohl2015deep,ho2020denoising}, which define $\Phi_\text{drift}(t, \z(t))\!=\!-\frac{1}{2}\beta(t)\,\z(t)$ and $\Phi_\text{diff}(t)\!=\!\sqrt{\beta(t)}$ for some function $\beta(t): [0,T]\rightarrow(0,1)$.

There are two common strategies to approximate the corresponding decoder, resulting in different change-of-variables formulas.
The first works by discretization of $\beta(t)$ into $0<\beta_1, ..., \beta_T<1$.
Then the encoder SDE turns into a Markov chain according to (\ref{eq:markov-chain-encoder}).
Using the special choices of the DDPM above, its conditionals $p(\Z_t\given\Z_{t-1})$ simplify into Gaussians
\begin{equation}
    p(\Z_t\given\Z_{t-1}\!=\!\z_{t-1}) = \normal\big(\Z_t;\,\sqrt{1-\beta_t}\,\z_{t-1},\, \beta_t\eye\big)
\end{equation}
The perturbation kernels, i.e. the conditionals at time point $t$ given the {\em input} $\Z_0=\x$, are also Gaussians
\begin{equation}\label{eq:sde-perturbation-kernel}
    p(\Z_t\given\Z_0=\x) = \normal\big(\Z_t;\,\sqrt{\alpha_t}\,\x, (1-\alpha_t)\eye\big)\qquad\text{with}\qquad\alpha_t=\prod_{\tau=1}^t(1-\beta_\tau)
\end{equation}
Since the $\alpha_t$ form a decreasing sequence with $\alpha_T\!\approx\! 0$, equation (\ref{eq:sde-perturbation-kernel}) shows that the encoder gradually transforms each data point into pure noise by pulling it towards the origin and simultaneously increasing the noise variance.
When the data distribution $p^*(\X)$ has unit variance, the DDPM parameterization preserves this property for all marginals $p_t(\Z_t)\!=\!\int p(\Z_t\given\Z_0=\x)\cdot p^*(\X\!=\!\x)\, d\x$.
Otherwise, it will gradually scale the variance to unity.
In particular, the induced code distribution $p_E(\Z_T)$, the last marginal in the chain, is arbitrarily close to standard normal as long as $\alpha_T$ is sufficiently small.
To learn the corresponding conditionals $p(\Z_{t-1}\given\Z_t)$ for the decoder chain (\ref{eq:markov-chain-decoder}), \autocite{ho2020denoising,song2021score} use the Gaussian parameterization
\begin{equation}
    p(\Z_{t-1}\given\Z_t\!=\!\z_t) = \normal\Big(\Z_{t-1};\, \frac{1}{\sqrt{1-\beta_t}}\big(z_t + \beta_t\,\s(\z_t, t)\big),\,\beta_t\eye\Big)
\end{equation}
where $\s(\z_t, t)$ is a learned vector field. 
Once these conditionals have been estimated, the probability of a generated sample can be calculated by the {\bf Markov chain CoV} (\ref{eq:markov-chain-cov}).

The second strategy discussed in \autocite{song2021score} reduces the decoder SDE to an ordinary differential equation (as described in \autoref{sec:neural-ode}) according to
\begin{equation}
    d\z = \big[\Phi_\text{drift}(t, \z(t)) - \frac{1}{2}\Phi_\text{diff}(t)^2\cdot\nabla_{\z} \log p_t(\z(t))\big]\, dt
\end{equation}
which is again executed backwards in time.
This ODE generates the same marginal data distribution $p(\X)\approx p^*(\X)$ as the original SDE, and its change-of-variables formula is the {\bf continuous CoV} (\ref{eq:continuous-cov}).
Flow matching \autocite{lipman2022flow} and rectified flows \autocite{liu2022flow} offer a very promising and efficient way of learning the vector field $F\big(t, \z(t)\big) := \big[\Phi_\text{drift}(t, \z(t)) - \frac{1}{2}\Phi_\text{diff}(t)^2\,\nabla_{\z} \log p_t(\z(t))\big]$ for this ODE in practice.

\subsection{Augmented Codes}

We speak of a model with augmented codes, when the code dimension is larger than the data dimension.
This necessarily results in a stochastic flow, because the model has at least $C>D$ independent variables.
An interesting instance of this idea are {\em augmented normalizing flows} \autocite{huang2020augmented}.
It defines a new variable $\XW=[\X, \Y]$ in data space by concatenating a noise vector with distribution $p^*(\Y)$ (typically standard normal) to the original data $\X$.
Since the noise is independent of the data, the true distribution of the augmented variable is the product of its constituents
\begin{equation}
    p^*(\XW) = p^*(\X)\cdot p^*(\Y)
\end{equation}
This distribution is now learned by a standard normalizing flow, resulting in the usual bijective change-of-variables formula
\begin{equation}
    p\big(\XW\!=\![\x,\y]\big) = p\big(\Z=f(\x,\y)\big) \, \big| \det \big(\J_f(\x,\y)\big)\big|
\end{equation}
Since we are only interested in the data part of $\xw=[\x, \y]$, we can in principle recover $p(\X\!=\!\x)$ as
\begin{equation}
    p(\X\!=\!\x) \overset{?}{=} \frac{p\big(\XW\!=\![\x,\y]\big)}{p^*(\Y\!=\!\y)} = p(\X\!=\!\x\given\Y\!=\!\y)
\end{equation}
However, in reality one does not achieve perfect convergence, so that the generated data distribution is not completely independent of the generated noise distribution, $p(\X\!=\!\x) \ne p(\X\!=\!\x\given\Y\!=\!\y)$.
To correct for this, \autocite{huang2020augmented} propose an importance sampling estimator, which approximates the marginalization integral $p(\X\!=\!\x) = \int p\big(\XW\!=\![\x,\y]\big)\, d\y$ as
\begin{flalign}\label{eq:augmented-flow-cov}
    \CoV{augmented flow CoV} &&  \nonumber \\
    && \hspace{-3cm}
    p(\X\!=\!\x) \approx \frac{1}{K} \nsum[1.3]_{k=1}^K \frac{p\big(\Z=f(\x,\y_k)\big) \, \big| \det \big(\J_f(\x,\y_k)\big)\big|}{p^*(\Y\!=\!\y_k)}\qquad\text{with}\qquad \y_k\sim p^*(\Y) &&
\end{flalign}
The idea behind augmented normalizing flows is that the enlarged data space facilitates the transformation of $p^*(\XW)$ to a standard normal code distribution $p(\Z)$.
However, since the intrinsic dimension of natural data tends to be much smaller than the ambient dimension, there is already plenty of spare space for the transformation to be learned accurately without augmentation.
It is therefore unclear under which conditions augmented normalizing flows are superior.



\section{Injective Flows}\label{sec:injective-flows}

The defining property of an injective flow is that the latent dimension is smaller than the data dimension, $\dim(\ZS) < \dim(\XS)$, that is, the codes form an information {\em bottleneck} \autocite{tishby2000information}.
In general, this results in a lossy encoding and encourages the encoder to focus on the core properties of the data.
In contrast to stochastic flows, which realize a bottleneck in terms of a stochastic encoder, this section discusses models with deterministic encoders.
An in-depth treatment of such architectures has been given in \autocite{nielsen2020survae} under the name of {\em SurVAE flows}, a term emphasizing the surjective nature of the encoder.
Interestingly, deterministic bottleneck encoders can be combined both with deterministic or stochastic decoders, giving rise to the fundamentally distinct ``injective flows'' (discussed here) and ``split flows'' (discussed in \autoref{sec:split-flows}), respectively.

The canonical example for an injective flow is the {\em autoencoder}.
Self-consistency of an autoencoder according to (\ref{eq:deterministic-self-consistency}) requires the decoder to be an injective function and the encoder its pseudo-inverse (precisely: left-inverse) surjective function.
Since we assume that $\X$ has full dimension $D$, injective flows with code dimension $C<D$ cannot define $p(\X)$ for every $\x \in \XS$, but only for the subset reachable by the decoder, the {\em decoder manifold}:
\begin{equation}
    \MS = \{g(\z)\in\XS: \z \in \ZS \}
\end{equation}
We abbreviate $\x \in \MS$ as $\x_\MS$.
The self-consistency requirement can now be expressed equivalently as
\begin{equation}
    \forall \x \in \MS:\quad \x_\MS \overset{!}{=} g(f(\x_\MS))
\end{equation}
In other words, points $\x_\MS$ are compressed without a loss, whereas all other $\x$ undergo lossy compression.
The equivalence classes of instances $\x$ mapped to the same code $\z$ are called the {\em fibers} of $\z$ resp. $\x$
\begin{equation}
    \FS(\z) = \{\x: \z = f(\x) \};\qquad\qquad \FS(\x) = \FS\big(f(\x)\big) 
\end{equation}
and we take $\x_\MS = g(f(\x))$ as the representative of the fiber $\FS(\x)$.
The union of the fibers $\bigcup_{x \in \MS} \FS(\x)$ spans the entire space $\XS$.
Since the encoder output is equal for points from the same fiber, we interpret fibers as the encoder's {\em nullspace}, in analogy to the nullspace of a linear projection.
Note that the pseudo-inverse represented by the encoder is not unique -- it must only be the decoder's exact inverse on the manifold.
Consequently, different encoders may result in different fiber geometries, even if the manifold is fixed.
The converse also holds: If the encoder is fixed, this determines the fibers, but the generator is still free to choose any point in a given fiber as its representative.

\subsection{Learning the Code Distribution}
\label{sec:code-distribution-learning}

For the autoencoder to be applicable as a generative model, one must make sure that the code distribution $p(\Z)$ is known.
This is not the case when the autoencoder is trained in the standard manner by minimizing some form of reconstruction loss, because this leads to an unknown and usually complicated induced code distribution $p_E(\Z) = f_\#p^*(\X)$.

There are two basic strategies to solve this problem: 
incorporate additional loss terms into the training objective, or learn $p_E(\Z)$ ex-post.
To realize the first strategy, \textcite{zhao2019infovae} propose to minimize the maximum mean discrepancy $\text{MMD}\big(p_E(\Z), p(\Z)\big)$ to pull $p_E(\Z)$ towards the desired $p(\Z)$, whereas \textcite{saha2022gens} minimize a kernel approximation of the Jensen-Shannon divergence, and \autocite{sorrenson2023maximum} use a linear combination of reconstruction error and maximum-likelihood loss. 
Perceptual generative autoencoders \autocite{zhang2020perceptual} add loss terms to enforce autoencoder self-consistency under both the prior and induced code distributions. 
Upon convergence, this implies $p_E(\Z) = p(\Z)$.
Invertible GANs \autocite{ghosh2022invgan} augment standard GANs such that the discriminator also adopts the role of an encoder, i.e. it uses an extra output head to transform data points $\x$ into corresponding codes $\z$, in addition to classifying them as reals or fakes.
The reconstruction loss in $\Z$-space now ensures that $p_E(\Z)$ converges to the GAN's prior $p(\Z)$.

The ex-post strategy works on top of conventionally trained autoencoders and adds a separate density model for $p_E(\Z)$, which then defines $p(\Z) := p_E(\Z)$.
Generative latent flows \autocite{xiao2019generative} and probabilistic autoencoders \autocite{bohm2020probabilistic} model $p_E(\Z)$ with a normalizing flow, which transforms the distribution $p_E(\Z)$ into a second latent space $\ZS'$ with $p(\Z')=\normal(0,\eye)$.
The effective encoder and decoder functions are now $\z'=\phi(f(\x))$ and $\x=g(\gamma(\z'))$, respectively, with $\gamma$ and $\phi=\gamma^{-1}$ representing the bijections learned by the normalizing flow.
Invertible interpretation networks \autocite{esser2020disentangling} apply the same architecture with the additional goal of making $\ZS'$ semantically disentangled.
Image generation with ``latent diffusion'' \autocite{rombach2022high} (subsequently re-implemented as ``stable diffusion'') uses a continuous normalizing flow, which is conditioned on the given user prompt to guide the code distribution towards the desired semantic content.
Other density estimators have been employed as well, for example Gaussians and Gaussian mixtures \autocite{ghosh2020variational}, energy models \autocite{pang2020learning,yoon2021autoencoding,vahdat2021score}, or discrete distributions after vector quantization of $\z$ as in VQ-VAEs \autocite{oord2017neural}.

\subsection{Autoencoders}\label{sec:autoencoders}

We assume in the sequel that the autoencoder has a known code distribution $p(\Z)$, because this is a prerequisite for it to possess a CoV formula.
To derive the resulting CoV formula, note that the volume change can no longer be expressed by the Jacobian determinant, since the Jacobian is now a rectangular matrix with undefined determinant (in contrast, Jacobians in the bijective case (\ref{eq:bijective-cov}) are square).
It turns out that this problem can be solved by calculating the determinant of the {\em squared} Jacobian (which has dimension $C\times C$) and equating the volume change with its square root.
Another crucial difference is that the autoencoder CoV formula does not represent the probability of individual data points $\x$, but of entire fibers $\FS(\x)$, because each code $\z$ collects the entire probability mass of the corresponding fiber $\FS(\z)$.
Thus one arrives at \autocite{krantz2008geometric,gemici2016normalizing} 
\begin{flalign}\label{eq:autoencoder-cov}
    \CoV{autoencoder CoV} &&\hspace{-1cm}
    p\big(\FS(\z)\big) = p\big(\x_\MS\!=\!g(\z)\big) = p\big(\Z\!=\!\z\big)\ \left|\det\left(\J_g^T \cdot \J_g\right)\right|^{-\frac{1}{2}} &&
\end{flalign}
For the sake of clarity, we have abbreviated the evaluation of the determinant at the current data point, $\J_g=\J_g(\z)$.
Intuitively, $p\big(\FS(\z)\big)$ is the density of a virtual dataset, where each point $\x$ has been shifted to the location of its representative $\x_\MS=g(f(\x))$.
Thus, formula (\ref{eq:autoencoder-cov}) defines a density {\em on the manifold} $\MS$, and not on the embedding space $\XS$, in line with the fact that the decoder has only $\dim(\ZS)$ degrees of freedom, and not $\dim(\XS)$ as would be required to fully model $\XS$.
\autoref{fig:donut-ae} demonstrated this behavior with a simple two-dimensional example.
In practice, one may interpret (\ref{eq:autoencoder-cov}) as an approximate density for $\x$ if all data points are very close to $\MS$, that is, when $\x \approx \x_\MS$. 
\autoref{sec:data-on-embedded-manifolds} describes specific model designs for data exactly located on a known manifold, and \autoref{sec:split-flows} discusses how equation (\ref{eq:autoencoder-cov}) can be extended to cover the entire space $\XS$ for data not on (or close to) $\MS$.

To understand why autoencoders cannot define a probability distribution for the entire space $\XS$, consider the hypothetical complementary CoV formula of the encoder
\begin{equation}\label{eq:hypthetical-encoder-cov}
    p(\X\!=\!\x)\, \overset{?}{=}\, p(\Z\!=\!f(\x))\ \left|\det\left(\J_f \cdot \J_f^T\right)\right|^{\frac{1}{2}}
\end{equation}
where again $\J_f = \J_f(\x)$ for clarity.
In the bijective case (\ref{eq:bijective-cov}), the encoder and decoder versions of the CoV formula are equivalent, but this only holds for points $\x_\MS\in\MS$ here.
Away from $\MS$, the code probability $p(\Z\!=\!f(\x))$ is constant for all points $\x \in \FS(\x)$ by definition, and the induced probability within the fiber varies only due to the changing determinant term.
In other words, this probability is solely determined by the geometry of $f(\x)$ and will almost certainly not reproduce the actual variation of the true distribution $p^*(\x\in\FS(\x))$ along the fiber, because this expression is not part of the reconstruction loss used to train $f(\x)$.

Various special cases have been studied to make the Jacobian determinant in (\ref{eq:autoencoder-cov}) more tractable. 
Linear models are considered in \textcite{cunningham2021change}.
Here, decoder and encoder are defined by a rectangular matrix $\W$ of size $D\times C$ and its pseudo-inverse $\W^+=(\W^T\W)^{-1}\W^T$:
\begin{equation}\label{eq:linear-autoencoder}
    g(\z) = \W \cdot \z; \qquad\qquad f(\x) = \W^+\! \cdot \x 
\end{equation}
The matrix $\W^+$ realizes a linear projection from $\XS$ to $\ZS$, and $\W$ maps $\ZS$ onto $\MS$, a $C$-dimensional linear subspace of $\XS$ corresponding to the image of $\W$.
Each fiber $\FS(\z)$ is normal to $\MS$ at $g(\z)$ and spans the nullspace of $\W^+$ in the classical linear algebra sense. 
The authors show that (\ref{eq:autoencoder-cov}) specializes into
\begin{flalign}\label{eq:linear-autoencoder-cov}
    \CoV{linear autoencoder CoV} &&\hspace{-14mm}
    p\big(\FS(\x)\big)  = p\big(\Z\!=\!\W^+\! \cdot \x\big)\ \left|\det\left(\W^T\! \cdot \W\right)\right|^{-\frac{1}{2}} &&
\end{flalign}
In principle, this formula turns dimension reduction with PCA into a generative model.
However, this is not so successful in practice because the underlying assumption of PCA -- that the data distribution is approximately Gaussian -- is rarely fulfilled.
Consequently, the code distribution $p(\Z)$ in (\ref{eq:linear-autoencoder-cov}) is usually complicated and must be learned by ex-post estimation of the induced code distribution $p_E(\Z)$, as described in \autoref{sec:code-distribution-learning}.
If one instead simply assumes a standard normal code distribution, the resulting generative model will not approximate $p^*(\X)$ well.

Specialized change-of-variables formulas for basic surjective operations 
-- tensor slicing, absolute value, ReLU, rounding, maximum/minimum, and sorting -- are derived in \autocite{nielsen2020survae}, along with the corresponding expressions for their inverses.
These formulas are primarily useful to define special layers in multi-layered architectures.
For example, $\MS$-flows \autocite{brehmer2020flows} and rectangular flows \autocite{caterini2021rectangular} apply padding (the inverse of slicing) in combination with two bijective functions to define the decoder and encoder as
\begin{equation}\label{eq:pad-slice}
    g(\z) = (g_D \circ \pad \circ\, g_C)(\z)\qquad\qquad f(\x) = (g_C^{-1} \circ \slice \circ\,g_D^{-1})(\x)
\end{equation}
where $g_D$ and $g_C$ are bijective in $\R^D$ and $\R^C$ respectively, $\text{pad}(\z') = [\z'; \mathbf{0}]$ concatenates a vector of $(D-C)$ zeros to the $C$-dimensional vector $\z'$, and $\text{slice}(\x')$ returns the first $C$ elements of vector $\x'$.
For points $\x_\MS$ on the generative manifold $\MS$, this results in the 
\begin{flalign}\label{eq:rectangular-flow-cov}
    \CoV{$\MS$-flow CoV} 
    && \hspace{-0.7cm}
    p\big(\x_\MS=g(\z)\big)  = p\big(\Z\!=\!\z\big)\ \left|\det\big(\J_{g_C}\big)\right|^{-1}\, \left|\det\big(\J^T_{\widetilde{g}_D}\cdot \J_{\widetilde{g}_D}\big)\right|^{-\frac{1}{2}} &&
\end{flalign}
Here, $\J_{\widetilde{g}_D}$ is the Jacobian of $(g_D \circ \text{pad})$, evaluated at $z'=g_C(\z)$.
The numerical computation of (\ref{eq:rectangular-flow-cov}) is potentially simpler than (\ref{eq:autoencoder-cov}).

\subsection{Autoencoders with Finite Codes}
\label{sec:finite-autoencoders}

The inability of autoencoders to generate the entire space $\XS$ becomes even more obvious when one considers finite codes $Z\in\{1,...,K\}$.
The $\FS(k)$ are now more appropriately called {\em facets} and have the same dimension as $\XS$.
They define a partition of $\XS$ such that $\XS=\bigcup_{k=1}^K \FS(k)$ and $\FS(k)\cap \FS(k')=\varnothing$ for $k\ne k'$.
Examples of popular facet types include the pixels of a grid, the Voronoi regions of a $K$-means clustering, and the bins of a histogram or density tree.
These and other classical methods are therefore injective flows in the sense of this section.
$\MS=\{\widehat{\x}_1,...,\widehat{\x}_K\}$ is no longer a manifold, but a finite set of representatives $\widehat{\x}_k$, for example the facets' geometric centroids or their centers of probability mass.
The encoder and decoder functions become
\begin{equation}
    f(\x) = \sum_{k=1}^K k\cdot\mathbbm{1}[\x\in\FS(k)]; \qquad\qquad g(k) = \widehat{\x}_k
\end{equation}
When facets are defined as Voronoi regions, data are assigned to the nearest representative with respect to some norm (usually Euclidean)
\begin{equation}
    \FS(k) := \big\{\x: \norm{\x - \widehat{\x}_k} < \norm{\x - \widehat{\x}_{k'}}\text{ for }k'\ne k \big\}\quad \Rightarrow \quad f(\x) = \argmin_k \norm{\x - \widehat{\x}_k}
\end{equation}
Self-consistency is guaranteed because $\widehat{\x}_k\in\FS(k)$.
The discrete latent probability can be obtained either exactly by integrating over the true distribution $p^*(\X)$ or approximately by counting the number of training points $\{\x_i\}_{i=1}^N$ in each facet
\begin{equation}\label{eq:facet-prior}
    p(Z=k)\,\, =\,\, \int_{\FS(k)} p^*(\X)\,d\X \,\,\approx\,\, \frac{1}{N} \sum_{i=1}^N \mathbbm{1}[\x_i \in \FS(k)]
\end{equation}
The CoV formula for the facets is a mixture of delta distributions, weighted by the code probabilities, so that sampling returns representatives $\widehat{\x}_k$ with their appropriate frequencies. 
We refer to this formula by the name ``$K$-means CoV'', but it equally applies to the other discretization methods mentioned.
\begin{flalign}\label{eq:finite-autoencoder-cov}
    \CoV{K-Means CoV} &&  \hspace{-2cm}
    p(\X\!=\!\x) =  \sum_{k=1}^K\, p(Z\!=\!k)\cdot\delta(\x - \widehat{\x}_k)&&
\end{flalign}
Since $k=f(\x)$ is constant within each facet, the Jacobian in (\ref{eq:hypthetical-encoder-cov}) is almost everywhere zero, and so will be $p(\X\!=\!\x)$ when $\x\notin\MS$.
Clearly, this is not an approximation of $p^*(\X)$ in most conceivable situations, showing again that autoencoders do not span the entirety of $\XS$. 

The performance of the $K$-means algorithm can be considerably improved when the clustering is applied {\em after} a non-linear transform of the data $\x$ into pre-codes $\widetilde{\z}=\phi(\x)$.
The term ``pre-codes'' is warranted, since the actual codes are the facet labels $1,...,K$.
This results in a {\em vector quantization} (VQ) autoencoder \autocite{oord2017neural, razavi2019generating}%
\footnote{The authors themselves call their model VQ-VAE, but in our terminology VAEs are stochastic flows, whereas VQ autoencoders are injective flows.
This exemplifies how our categorization can clarify subtle differences between methods.}.
The facets now form a partition of the pre-code space $\widetilde{\ZS}=\bigcup_{k=1}^K \FS(k)$ with representatives $\widehat{\z}_1,...,\widehat{\z}_K$.
Encoder and decoder are thus defined as
\begin{equation}
    f(\x) = \sum_{k=1}^K k\cdot\mathbbm{1}[\widetilde{\z}=\phi(\x)\in\FS(k)]; \qquad\qquad \widehat{\x}_k=g(k) = \gamma(\widehat{\z}_k)
\end{equation}
where the function $\gamma(\widehat{\z}_k)$ maps representatives from pre-code space to data space.
Self-consistency requires $k=f(g(k))$.
After convergence of the encoder, the prior distribution $p(Z=k)$ of the codes is learned by counting as in equation (\ref{eq:facet-prior}), but with the facets $\FS(k)$ now defined in pre-code space: $f(\x)=\argmin_k \norm{\phi(\x) - \widehat{\z}_k}$.
The probability mass associated with each decoder output $\widehat{\x}_k=\gamma(\widehat{\z}_k)$ is still defined by equation (\ref{eq:finite-autoencoder-cov}). 

VQ autoencoders solve a major problem of standard $K$-means: 
Since $K$-means defines the data space partition in terms of the Voronoi regions of the representatives, it requires a metric that groups data instances according to their semantic similarity.
In most applications, no such metric is known, and one instead resorts to standard metrics (e.g., Euclidean distance) that often fail to induce meaningful groupings.
In contrast, learned pre-codes re-arrange the data such that the Euclidean distance {\em becomes} semantically meaningful in the embedding space, resulting in a more useful assignment of data instances to representatives.

To achieve diversity in the decoder outputs for complex data types like images, the number of codes $K$ must be very large.
VQ autoencoders maintain tractability in the image setting by using relatively small (e.g. with $K_0=512$) sets of {\em local} codes, and representing original inputs as low resolution {\em arrays} of local codes (with size $w\times h=32\times32\,...\,128\times128$, depending on the original image size).
The effective code size is then $K=K_0\cdot w\cdot h$, and the prior $p(Z)$ is decomposed into a joint distribution over local code arrays
\begin{equation}
    p(Z) = p(Z_{11},...,Z_{wh})
\end{equation}
This distribution is then learned by an auto-regressive network, such as PixelCNN \autocite{oord2016conditional}.

To ensure diversity with much fewer codes, distribution-preserving lossy compression (DPLC) proposes an elegant {\em stochastic} decoder design \autocite{tschannen2018deep,blau2019rethinking}. 
In its simplest form, the method assumes that codes are located on a finite regular grid, $\ZS\subset \integers^C$. 
Then the decoder can be trained to generate a distribution for the entire facet $p\big(\X\!=\!\x\given\FS(\z)\big)$ from {\em noisy} codes $\z'=\z+\s$ with $\s\sim\uniform(-0.5, 0.5)^C$, in addition to returning good representatives $\widehat{\x}_{\z}$ for unperturbed codes $\z$.
Note that the noise is chosen such that the $\z'$ never leak into neighboring facets.
Results of this method with $K=2^2,..., 2^{12}$ are shown in \autoref{fig:mnist-rate-distortion-perception-trade-off} in \autoref{sec:coding-theory}.
The resulting architecture resembles a split flow, but still is an injective flow, because the dimension of the decoder manifold $\MS$ cannot exceed the dimension of the noise $\s$.


 
\subsection{Data on Embedded Manifolds}
\label{sec:data-on-embedded-manifolds}

When the data are located on a $C$-dimensional manifold $\MS$ embedded in the $D$-dimensional ambient space $\XS$, the dimension mismatch excludes the possibility of a bijective mapping, i.e. standard normalizing flows are not applicable.
However, a bottleneck architecture with $\dim(\ZS)=\dim(\MS)$ can represent all data points exactly {\em without} losing information.
This is easiest when $\MS$ is a known geometric object equipped with a known bijective map $\x=\phi(\widetilde{\z})$ from a single $C$-dimensional {\em chart} $\widetilde{\ZS}$, possibly excluding a zero set of the data (under Lebesgue measure on $\MS$, e.g., one pole of a sphere).
Since $\dim(\ZS)=\dim(\MS)$, a bijective flow (e.g., a normalizing flow) can now learn the distribution on the chart, which is then be mapped back onto $\MS$ by $\phi$.

This is the idea behind {\em manifold flows} \autocite{gemici2016normalizing,rezende2020normalizing}.
They express the target distribution $p(\X)$ as a pushforward through $\phi$ of a corresponding distribution $p(\widetilde{\Z})$ on the chart:
\begin{equation}
    p(\X\!=\!\x) = \phi_\#p\big(\widetilde{\Z}\big) = 
    p\big(\widetilde{\Z}=\phi^+(\x)\big)\cdot\big|\det\!\big(\J^T_{\phi}\ \J_{\phi}\big)\big|^{-\frac{1}{2}} 
\end{equation}
where $\phi^+(\x)$ denotes the left-inverse of $\phi(\widetilde{\z})$ (that is, the mapping from the manifold onto the chart), and $\J_{\phi}$ is the Jacobian of $\phi$ at $\widetilde{\z}=\phi^+(\x)$.
Now, a standard normalizing flow $f(\widetilde{\z})$ is trained to map from chart space to code space:
\begin{equation}
     p\big(\widetilde{\Z}=\widetilde{\z}\big) = p\big(\Z=f(\widetilde{\z})\big)\cdot \big|\det\!\big(\J_f(\widetilde{\z})\big)\big|
\end{equation}
Combining both equations gives the manifold flow change-of-variables formula
\begin{flalign}\label{eq:manifold-flow-cov}
    \CoV{manifold flow CoV} &&  \nonumber \\
    && \hspace{-2.8cm}
    p(\X\!=\!\x) = p\big(\Z\!=\!f(\phi^+(\x))\big) \cdot 
    \big|\det\!\big(\J^T_{\phi}\ \J_{\phi}\big)\big|^{-\frac{1}{2}} \!\cdot 
    \big|\det\!\big(\J_f(\phi^+(\x))\big)\big|  && 
\end{flalign}
Recall that $\phi$ and $\J_\phi$ are analytically known, and only $f$ and $\J_f$ must be learned.
We note in passing that some authors \autocite{mathieu2020riemannian,falorsi2020neural} have reported competitive or even superior performance in a similar setting by replacing the projection method with {\em continuous} normalizing flows for Riemannian manifolds.

In addition, \textcite{rezende2020normalizing} propose an alternative possibility to define the code distribution $p(\Z)$, which is applicable when $\MS$ is compact and has finite volume $V_\MS$, such as a (hyper-) sphere or torus.
Then, $p(\Z)$ can be defined as the pushforward of the uniform distribution on $\MS$ through $\phi^+$:
\begin{equation}
    p(\Z\!=\!\z) = {\phi^+}_{\!\#}\text{Uniform}(\MS) = \frac{1}{V_\MS} \cdot \big| \det\!\big(\J^T_{\phi}(\z)\ \J_{\phi}(\z)\big)\big|^\frac{1}{2} 
\end{equation}
This expression replaces $p(\Z)$ in equation (\ref{eq:manifold-flow-cov}) and avoids complications that arise because a standard normal $p(\Z)$ is incompatible with a chart of finite support.

When the manifold is unknown, the mapping $\phi$ from the chart to $\MS$ must be learned in addition to the normalizing flow $f$.
{\em Conformal embedding} flows \autocite{ross2021tractable} restrict $\phi$ to (piece-wise) conformal mappings, which means that $\J^T_{\phi}(\widetilde{\z})\ \J_{\phi}(\widetilde{\z}) = \lambda(\widetilde{\z})^2\cdot \eye_C$ for some scalar function $\lambda(\widetilde{\z})$ -- a restriction acting as a strong regularizer.
The Jacobian of the projection thus simplifies, resulting in the change-of-variables formula
\begin{flalign}\label{eq:conformal-embedding-cov}
    \CoV{conformal embedding CoV} && \nonumber \\
    && \hspace{-4cm}
    p(\X\!=\!\x) = p\big(\Z=f(\phi^+(\x))\big)
    \cdot \lambda\big(\phi^+(\x)\big)^{-C}\!\! \cdot  \big|\det\!\big(\J_f(\phi^+(\x))\big)\big| &&
\end{flalign}
Since globally conformal maps are too restrictive for most problems of interest, \autocite{ross2021tractable} propose to relax this requirement to {\em piece-wise conformal} mappings.
An especially interesting realization of this idea are the {\em VQ flows} \autocite{sidheekh2022vq}.
Here, $\MS$ is represented not by a single chart, but by an {\em atlas} (i.e. a collection of charts).
To decide about the chart responsible for each point $\x$, they take the same approach as VQ autoencoders:  
An additional encoder $\psi^+(\x)$ maps the data to some feature space, where chart membership is determined by the nearest representative: $h(\x)=\argmin_k\norm{\psi^+(x)-\widehat{\z}_k}$.
Thus, $h(\x)$ returns the chart label, and a separate conformal flow is learned for each chart:
\begin{flalign}\label{eq:conformal-vq-flow-cov}
    \CoV{conformal VQ flow CoV} && \\
    && \hspace{-4.0cm}
    p(\X\!=\!\x) = \sum_{k=1}^K\,\mathbbm{1}[h(\x)\!=\!k]\cdot p(k)\cdot
    p\big(\Z\!=\!f_k(\phi^+_k(\x))\big) \cdot 
    \lambda_k\big(\phi^+_k(\x)\big)^{-C}\!\! \cdot \big| \det\!\big(\J_{f_k}(\phi^+_k(\x)))\big)\big|    &&  \nonumber
\end{flalign}
Since the charts can be made arbitrarily small, each conformal transformation $\phi_k$ is only responsible for a small part of $\MS$'s geometry, which alleviates the limitations of global conformal maps.
\textcite{sidheekh2022vq} report very good results for 2-dimensional manifolds embedded in 3D, but it is as yet unclear how the method scales to higher dimensions \autocite[see the discussion in][]{alberti2023manifold}.

Standard normalizing flows cannot exactly represent distributions on embedded manifolds, because the dimension mismatch prevents the map from being bijective.
However, they can learn such distributions approximately with high accuracy using the {\em SoftFlow} approach \autocite{kim2020softflow} or, equivalently, {\em conditional denoising NFs} \autocite{loaiza2022denoising}.
Instead of learning $p(\x\in\MS)$, these methods learn a model for the noisy data $p(\x + \epsilon)$, where $\epsilon\sim\normal(0, \sigma^2 \cdot \eye_D)$ is Gaussian.
Since $\x + \epsilon$ is a D-dimensional set, it can be learned by a normalizing flow.
Crucially, this normalizing flow is {\em conditioned} on the noise level $\sigma$, and the noise level is varied during training according to some prior distribution, such as $\sigma\sim\text{Uniform}(\sigma_\text{min}, \sigma_\text{max})$.
In this way, the conditional normalizing flow learns how the spread of the data around $\MS$ varies in accordance to the given noise level $\sigma$.
To generate data approximately on $\MS$, the converged model is executed at noise level $\sigma_\text{min}$, giving the change-of-variables formula
\begin{flalign}\label{eq:softflow-cov}
    \CoV{SoftFlow CoV} && \hspace{-0.0cm}
    p(\x_\MS\!\in\!\MS) \approx 
    p\big(\Z\!=\!f(\x; \sigma_\text{min})\big) \cdot 
    \big|\det\!\big(\J_f(\x; \sigma_\text{min})\big)\big|    && 
\end{flalign}
However, experiments in \autocite{loaiza2022denoising} suggest that $\sigma_\text{min}$ cannot be made as small as one would like -- the performance of the model saturates at a certain lower bound and does not improve further by decreasing $\sigma_\text{min}$.


\todo{discuss \autocite{voleti2021multi,altekruger2023neural}, Schrödinger bridges?}

\section{Split Flows}\label{sec:split-flows}

In vanilla autoencoders, both encoder and decoder are deterministic, whereas in stochastic flows both are probabilistic.
We now consider a third possibility, which we call {\em split flows}: to combine {\em deterministic} encoders with {\em stochastic} decoders.
The idea is to extend the autoencoder CoV formula (\ref{eq:autoencoder-cov}) to the entire space $\XS$ (i.e. beyond the decompression set/manifold $\MS \subset \XS$) by learning a conditional distribution $p\big(\X\given\FS(\z)\big)$ for the residual behavior of $p^*(\X)$ {\em within} the fiber $\FS(\z)=\FS(f(\x))$. 
We can then define the CoV formula abstractly as
\begin{flalign}\label{eq:split-nf-cov}
    \CoV{split flow CoV} && \hspace{-2cm}
    p(\X\!=\!\x) = p\big(\FS(f(\x))\big)\cdot p\big(\X\!=\!\x\given\FS(f(\x))\big) &&
\end{flalign}
This is essentially an infinite mixture model with mixture components $p\big(\X\!=\!\x\given\x\in\FS(\z)\big)$ and mixture weights $p\big(\FS(\z)\big)$.
This mixture does {\em not} define a stochastic flow, because fibers are pairwise disjoint and $p\big(\X\!=\!\x'\given\x'\notin\FS(\z)\big) := 0$ by construction.
Thus, each point $\x$ is only influenced by a single code $\z=f(\x)$, and the stochastic decoder defined by the RHS of (\ref{eq:split-nf-cov}) can be self-consistently combined with the {\em deterministic} encoder defined by $f(\x)$.
In contrast, if a conditional $p\big(\X\!=\!\x'\given\FS(\z)\big)$ placed mass outside of $\FS(\z)$, some or all $\x'$ would be influenced by multiple codes, and a stochastic encoder $p(\Z\given\X\!=\!\x')$ would be required to model the code ambiguity in a self-consistent manner.

Like a bijective flow, a split flow exactly spans the space $\XS$.
However, it separates the representation into a bottleneck code of dimension $C = \dim(\ZS) < \dim(\XS) = D$ that only models $p\big(\FS(\x)\big)$ on the manifold $\MS$, and a conditional $p\big(\X\!=\!\x\given\FS(\x)\big)$ representing the remaining $D-C$ independent dimensions {\em off} the manifold (i.e., in $\MS$'s nullspace).
The separation into essential (``core'') behavior on $\MS$ and less important (``detail'') behavior in $\FS$ promises to improve model robustness and interpretability in comparison to standard NF representations of the same data.

The product of the two distributions of a split flow has $D$ independent dimensions, and its joint distribution $p(\X, \Z)$ is degenerate due to $\z=f(\x)$ being deterministic.
This differs from a stochastic flow with non-degenerate $p(\X, \Z)$, which has $D+C$ independent dimensions, and from an injective flow, whose joint distribution $p(\X, \Z)$ is even more degenerate (namely restricted to $\MS$) with only $C$ independent dimensions.

{\em Normalized autoencoders} \autocite{yoon2021autoencoding} are a direct realization of (\ref{eq:split-nf-cov}), although the paper does not present the method in this way.
The authors propose to implement the fiber conditionals by a Gibbs distribution over the squared distances between data points and their reconstructions:
\begin{equation}
    p\big(\X\!=\!\x\given\x\in\FS(\widehat{\x})\big) = \frac{1}{B}\exp\big(-\norm{\x - \widehat{\x}}^2/T\big)\cdot\mathbbm{1}[\x\in\FS(\widehat{\x})]
\end{equation}
where $B$ is the normalization constant and $T$ the temperature.
This is equivalent to the assumption that the data density in the fibers is a Gaussian centered on $\MS$.
Since this distribution is postulated, not learned, their training algorithm acts on the encoder function $f(\x)$ and manipulates the {\em fiber geometry}, that is, it changes where the representatives $\widehat{\x}$ are placed and which points are projected onto each representative.
Samples from their method look promising, but this is preliminary because the focus of the paper is outlier detection, not generative modeling. 

\todo{discuss approximation by cINNs, e.g. Felix' experiments, feature analysis by \autocite{esser2020disentangling,rombach2020making}}

\subsection{Discrete and Linear Split Flows}
\label{sec:discrete-and-linear-split-flows}

We get the simplest incarnation of split flows in a setting with discrete codes $Z\in\{1,...,K\}$ (e.g. rounding to a finite grid, histograms/density trees, or $K$-means clustering, see section \ref{sec:injective-flows}) by distributing the code probability $p(Z=k)$ uniformly over the facets $\FS(k)$.
This results in $p\big(\X\!=\!\x\given\FS(f(\x))\big) = 1/|\FS(f(\x))|$, where $|\FS(f(\x))|$ is the facet's volume (recall that $f(\x)$ computes the index $k$ of the facet containing $\x$).
Of course, we must make sure that the facets have finite volume, e.g. by bounding them to the convex hull of the training data.
The decoder then consists of two steps: first, sample an index $z \sim p(Z)$, and then sample a point uniformly from the facet $\FS(z)$.
In extension of equation (\ref{eq:finite-autoencoder-cov}), the corresponding CoV formula becomes
\begin{flalign}
    \CoV{piecewise constant CoV} && \hspace{-24mm}
    p(\X\!=\!\x) = p(Z\!=\!f(\x))\cdot \frac{1}{|\FS(f(\x))|} &&
\end{flalign}
It is well known that this formula converges to $p^*(\X)$ in the asymptotic limit $K\rightarrow\infty$.
Better approximations of $p^*(\X)$ with discrete codes are achievable with more sophisticated models for the facet conditionals $p\big(\X\!=\!\x\given\FS(\z)\big)$ in (\ref{eq:split-nf-cov}).
A successful heuristic defines these conditionals by fitting a Gaussian to the true distribution $p^*(\X)$ within each facet \autocite{criminisi2013decision}.
Accurate conditionals in terms of piece-wise normalizing flows (one NF per facet) are obtained by VQ flows \autocite{sidheekh2022vq}, which were discussed in \autoref{sec:finite-nfs} and \autoref{sec:data-on-embedded-manifolds}.

Another instructive case is the extension of the linear autoencoder (\ref{eq:linear-autoencoder}) to the entire space $\XS$ \autocite{cunningham2021change}.
Recall that the decoder is defined by $\x=\W\cdot\z$ with matrix $\W$ and $C$-dimensional code $\z$, and a self-consistent encoder is obtained via the pseudo-inverse $\W^+$ as $\z=\W^+\cdot\x$.
Then, $\MS$ is a $C$-dimensional linear subspace of $\XS$, and the fibers are orthogonal to $\MS$.
Let the singular value decomposition of $\W$ be
\begin{equation}
    \W = \begin{bmatrix} \U_\parallel & \U_\bot \end{bmatrix}
    \begin{bmatrix} \mathbf{\Lambda} \\ \0 \end{bmatrix} \V^T
\end{equation}
$\U_\parallel$ and $\U_\bot$ form an orthonormal basis for $\MS$ and for the fibers respectively.
We can thus define a complementary $(D-C)$-dimensional code $\z_\bot$ for the fibers (i.e., the encoder's nullspace), by 
\begin{equation}
    \z_\bot = \U^T_\bot\cdot \x
\end{equation}
The difference vector between a point $\x\in\XS$ and its representative $\x_\MS\in\MS$ can be expressed by
\begin{equation}
    \x - \x_\MS = \x -\W \W^+\!\cdot \x = \U_\bot\cdot \z_\bot
\end{equation}
Defining a nullspace conditional $p(\Z_\bot \given \Z)$, the pushforward of the linear split flow decoder becomes
\begin{equation}\label{eq:linear-stochastic-nullspace-pushforward}
    \x\sim p(\X)\quad\Longleftrightarrow\quad \x = \W \cdot \z + \U_\bot \cdot\z_\bot\quad\text{ with }\,\z\sim p(\Z)\, \text{ and }\,\z_\bot \sim p(\Z_\bot \given \Z\!=\!\z)
\end{equation}
Consequently, \textcite{cunningham2021change} arrive at the following change-of-variables formula, which is an extension of (\ref{eq:linear-autoencoder-cov}) and a special case of (\ref{eq:split-nf-cov})
\begin{flalign}\label{eq:linear-stochastic-nullspae-cov}
    \CoV{linear split flow CoV} && \nonumber \\
    &&  \hspace{-40mm} p(\X\!=\!\x) = p\big(\Z\!=\!\W^+\!\cdot\x\big)
    \cdot \left|\,\det\left(\W^T\! \cdot \W\right)\right|^{-\frac{1}{2}}
    \cdot p\big(\Z_\bot\!=\!\U^T_\bot\cdot\x\,\big|\,\Z\!=\!\W^+\!\cdot\x\big) &&
\end{flalign}
Note that the nullspace conditional does not require an additional determinant term because $\U_\bot$ is orthonormal and therefore $\det(\U^T_\bot\!\cdot\U_\bot)=1$.

\subsection{Split Flows with Predefined Encoders}

A universal strategy to generalize these principles to non-linear encoders is to apply the linear projection {\em after} a non-linear bijective transformation $\varphi$, resulting in the encoder:
\begin{equation}
    \z = \W^+ \cdot \varphi(\x)
\end{equation}
The projection $\W^+$ can now be defined in the simplest possible way -- it just extracts the first $C$ output dimensions of the non-linear function $\varphi$, i.e. it is equal to the $\slice$ operation in (\ref{eq:pad-slice}).
The complementary projection for the encoder's nullspace selects the remaining dimensions of $\varphi$'s output:
\begin{equation}
    \z = \begin{bmatrix} \eye_C & \0 \end{bmatrix} \cdot \varphi(\x) =: \slice \cdot\, \varphi(\x)
    \qquad\qquad
    \z_\bot = \begin{bmatrix} \0 & \eye_{D-C} \end{bmatrix} \cdot \varphi(\x) =: \slice_\bot \cdot\, \varphi(\x)
\end{equation}
Any more complicated $\W^+$ can be equivalently absorbed into $\varphi$, provided that $\varphi$ is taken from a sufficiently expressive function family.
The simple choice of $\W^+=\slice$ also ensures that $\det(\slice\cdot\slice^T)=1$, so that no extra determinant term needs to be inserted into the CoV formula.

A vivid example of this technique is described in \textcite{ardizzone2020conditional} for gray-scale image  colorization.
Here, the encoder consists of a pixel-wise non-linear transformation $\varphi$ from RGB color space (where brightness and color are entangled) to Lab color space (where brightness `L' and color `ab' are independent), followed by dropping the color channels via multiplication with $\slice$.
Thus, the code and its complement are $\z =\text{L}$ and $\z_\bot=\text{ab}$, with corresponding random variables $\Z_\text{L}$ and $\Z_\text{ab}$.
The encoder's fibers $\FS(\z)$ consist of all color images sharing the same gray-scale channel $\z$, and the decoder has the task of generating plausible (i.e., highly probable) color images within $\FS(\z)$. 

The authors solved the above problem by supervised training of a conditional normalizing flow for $p(\Z_\text{ab}\given\Z_\text{L})$, resulting in the decoder
\begin{equation}
    \x = \varphi^{-1}\left(\begin{bmatrix} \z \\ \z_\bot \end{bmatrix}\right) \quad\text{ with }\,\z=\slice\cdot\,\varphi(\x)\,\text{ and }\,\z_\bot\sim p(\Z_\text{ab}\given\Z_\text{L}\!=\!\z)
\end{equation}
where $\varphi^{-1}$ transforms Lab back into RGB. 
This is clearly a split flow, because the conditional generator cannot leak out of the facet $\FS(\z)$ by construction.
The corresponding CoV formula is again a special case of (\ref{eq:split-nf-cov}), namely the product of three terms: (i) a generative distribution $p(\Z_\text{L})$ for gray-scale images, (ii) the conditional $p(\Z_\text{ab}\given\Z_\text{L})$ expanded according to the CoV formula for conditional NFs (\ref{eq:condnf-cov}), and (iii) the Jacobian determinant of $\varphi$.
It has the same functional form as formula (\ref{eq:split-normalizing-flow-cov}) below.

A similar idea is the basis of {\em wavelet flows} \autocite{yu2020wavelet} and {\em invertible image rescaling} \autocite{xiao2020invertible}:
An orthogonal wavelet transform $\varphi$ decomposes a given image $\x$ into a lower resolution version $\z$ of itself and three channels $\z_\bot$ of detail coefficients.
By learning a conditional NF for the conditional $p(\Z_\text{details}\given\Z_\text{low-res}\!=\!\z)$, the decoder can generate a diverse set of high-quality upsampling solutions for the same low-res image $\z$.

\subsection{Learning the Split in Normalizing Flows}
\label{sec:split-nfs}

While the colorization and upsampling examples use predefined transformations $\varphi$ (e.g., RGB to Lab conversion, wavelet transforms), various authors have generalized this idea to $\varphi$ that are learned as finite or continuous normalizing flows.
We call this model class {\em split normalizing flows}, because the split flow idea is realized by a normalizing flow whose code space $\ZS$ is split into a $C$-dimensional {\em core} part $\ZS_c$ and a $(D-C)$-dimensional {\em detail} part $\ZS_d$.
The two subspaces define a decoder manifold $\MS$ and fibers $\FS(\z_c)$ by the following equivalences
\begin{align}
    \x\in \MS & \qquad\Longleftrightarrow\qquad \x = \varphi^{-1}\left(\begin{bmatrix} \z_c \\ \0 \end{bmatrix}\right)
    \quad \text{with } \z_c\sim p(\Z_c) 
    \label{eq:split-nf-manifold}\\
    \x \in \FS(\z_c) & \qquad\Longleftrightarrow\qquad \x = \varphi^{-1}\left(\begin{bmatrix} \z_c \\ \z_d \end{bmatrix}\right)
    \quad \text{with } \z_c\text{ fixed and } \z_d\sim p(\Z_d\given\Z_c\!=\!\z_c)
    \label{eq:split-nf-fiber}
\end{align}
It is easy to see that (\ref{eq:linear-stochastic-nullspace-pushforward}) is a special case of these formulas with linear $\varphi^{-1}$.
The crucial property of these definitions is that the bijectivity of $\varphi$ guarantees that the pushforward (\ref{eq:split-nf-fiber}) for one fiber can never leak to points in another fiber.
In other words, split normalizing flows are split flows by construction.
The deterministic encoder of a split NF is defined as
\begin{equation}
    \z_c = \slice \cdot\, \varphi(\x) =: \varphi_c(\x) \qquad\quad
    \z_d = \slice_\bot\cdot\, \varphi(\x) =: \varphi_d(\x)
\end{equation}
with $\slice$ again selecting the first $C$ dimensions.
The simplest way to learn a split that actually puts the essential information into $\z_c$ and leaves the details to $\z_d$ is to train the normalizing flow like an autoencoder \autocite{nguyen2019training,brehmer2020flows} by minimizing the expected $L_2$ reconstruction error of the core variable alone
\begin{equation}\label{eq:split-normalizing-flow-loss}
    \mathcal{L}[\varphi] = \E_{p^*(\X)}\!\left[\norm{\x - \x_\MS}^2_2\right]\quad
    \text{with}\quad \x_\MS = \varphi^{-1}\left(\begin{bmatrix} \varphi_c(\x) \\ \0 \end{bmatrix}\right)
\end{equation}
Let $\widehat{\varphi} = \argmin_\varphi \mathcal{L}[\varphi]$ be the solution of this learning problem.
The induced encoder pushforward
\begin{equation}
    p_E(\Z) = p_E\left(\begin{bmatrix} \Z_c \\ \Z_d \end{bmatrix}\right) = \widehat{\varphi}_\# p^*(\X)
\end{equation}
is in general a complicated distribution, because the loss (\ref{eq:split-normalizing-flow-loss}) does not place any constraints on $p_E(\Z)$, in contrast to standard normalizing flow training.
In their $\MS$-flow approach, \textcite{brehmer2020flows} suggest to express the core part $p_E(\Z_c)$ by a second normalizing flow $\s_c=\psi_c(\z_c)$ with outsourced noise distribution $p(\S_c)$.
This results in the same model as the probabilistic autoencoder \autocite[, see \autoref{sec:injective-flows}]{bohm2020probabilistic}, except that the autoencoder is now implemented by a NF with suppressed details $\z_d=\0$.
Its change-of-variables formula is identical to (\ref{eq:autoencoder-cov}), where the Jacobian $\J_g$ is defined as the $(D\times C)$ matrix 
$\J_g = \partial \x_\MS \big/ \partial \z_c$.

\todo{mention \autocite{cunningham2020normalizing,cramer2022nonlinear,khorashadizadeh2023conditional}}

However, $\MS$-flows are not split flows but injective flows, because the data distribution {\em within} the fibers is not learned.
We can easily turn them into split flows by expanding the full encoder pushforward $p_E(\Z) = p_E(\Z_c)\cdot p_E(\Z_d\given\Z_c)$ according to Bayes rule.
This requires to represent $p_E(\Z_d\given\Z_c)$ by yet another normalizing flow (this time a conditional one), $\s_d=\psi_d(\z_d, \z_c)$, with outsourced noise distribution $p(\S_d)$.
Then, sampling from $p_E(\Z)$ is realized by the pushforwards
\begin{align}
    \z_c\sim p_E(\Z_c) & \qquad\Longleftrightarrow\qquad 
    \z_c = \psi^{-1}_c(\s_c)
    &\text{with}\quad  \s_c\sim p(\S_c) 
    \label{eq:split-nf-core} \\
    \z_d \sim p_E(\Z_d\given \Z_c\!=\!\z_c) & \qquad\Longleftrightarrow\qquad 
    \z_d = \psi^{-1}_d(\s_d, \z_c)
    &\text{with}\quad  \s_d\sim p(\S_d)
    \label{eq:split-nf-detail}
\end{align}
Putting everything together, we finally arrive at the change-of-variables formula specializing (\ref{eq:split-nf-cov}) into  
\begin{align}
    \hspace{-10mm}\CoV{split normalizing flow CoV} \nonumber \\
    p(\X\!=\!\x) = &\, 
       \big|\det\big(\J_{\widehat{\varphi}}(\x)\big)\big| \cdot 
       p_E\big(\Z_c\!=\!\widehat{\varphi}_c(\x)\big)\cdot
       p_E\big(\Z_d\!=\!\widehat{\varphi}_d(\x)\,\big|\,\Z_c\!=\!\widehat{\varphi}_c(\x)\big) \label{eq:split-normalizing-flow-cov}  \\[3mm]
       = &\, \big|\det\big(\J_{\widehat{\varphi}}(\x)\big)\big| \cdot 
       \big|\det\big(\J_{\psi_c}\big(\widehat{\varphi}_c(\x)\big)\big)\big| \cdot
       \big|\det\big(\J_{\psi_d}\big(\widehat{\varphi}_d(\x), \widehat{\varphi}_c(\x)\big)\big)\big| \nonumber \\[1.5mm]
       & \cdot p\big(\S_c\!=\!\psi_c\big(\widehat{\varphi}_c(\x)\big)\big)
       \cdot p\big(\S_d\!=\!\psi_d\big(\widehat{\varphi}_d(\x), \widehat{\varphi}_c(\x)\big)\big) \label{eq:split-normalizing-flow-cov-expanded}
\end{align}
where we have expanded the code probabilities according to (\ref{eq:bijective-cov}) and (\ref{eq:condnf-cov}) in the second equation.
The two-stage training process -- first learn $\widehat{\varphi}$ by minimizing the reconstruction loss (\ref{eq:split-normalizing-flow-loss}), then learn $\psi_c$ and $\psi_d$ by maximizing the data likelihood (\ref{eq:split-normalizing-flow-cov-expanded}) -- works well in the $\MS$-flow experiments of \textcite{brehmer2020flows}, albeit without learning $\psi_d$.
Moreover, some recent results suggest that improved optimization methods will also allow joint learning of $\varphi$ and $\psi_c$ \autocite{caterini2021rectangular}.

An alternative training strategy is put forward by {\em denoising normalizing flows} \autocite{horvat2021denoising}, which use a simplified version of equations (\ref{eq:split-nf-core}) and (\ref{eq:split-nf-detail}) by taking $\psi_d$ as the identity function, thereby eliminating $|\det(\J_{\psi_d})|$ from (\ref{eq:split-normalizing-flow-cov-expanded}).
The authors make the crucial assumption that the data are concentrated near an embedded manifold $\MS$, which means that the data variability {\em within} the manifold is much bigger than perpendicular to it.
In other words, there exists a range of variances $[\sigma_1^2, \sigma_2^2]$ such that $\sigma_1^2$ is greater than the off-manifold data variance, but $\sigma_2^2$ is still considerably less than the within-manifold variance.
When the data is now augmented with Gaussian noise between $\sigma_1^2$ and $\sigma_2^2$, the noise will not markedly change the observed variance within $\MS$, but will significantly increase it in $\MS$'s nullspace.
Consequently, they detect the core dimensions by being insensitive to the added noise, whereas the details are influenced by the noise.
This is realized by a training objective that combines the reconstruction loss of the $\MS$-flow on the core dimensions with a maximum likelihood loss on the entire code.
Experiments show good separation between core and detail on challenging datasets like the 2-D spiral or StyleGAN generated images.

The realization of split flows by normalizing flows whose code space is split into core and detail dimensions highlights the dual nature of the method:
When we consider core and details together, the model behaves like a standard normalizing flow with a lossless bijective encoding.
However, when we zero out the detail part of the code and preserve only the core, we get a lossy surjective encoding.
Sampling new details from their latent distribution then gives a reconstruction that preserves the essence of the original data instance, but differs in the details.

\subsection{Disentangled Normalizing Flows}

The introduction of the additional NFs $\psi_c$ and $\psi_d$ can be avoided when $p_E(\Z)$ is directly forced to conform to a tractable known distribution $p(\Z)$, as in standard NF training.
For clarity, we return to the simpler original notation $g(\z)=\varphi^{-1}(\z)$ and $f(\x)=\varphi(\x)$ in this setting.
Various algorithms of this kind have been proposed in the context of {\em latent disentanglement}, which aims to identify codes whose individual dimensions are associated with distinct and interpretable changes in the data space.
The set of {\em important} (by some definition) code dimensions will then form the core subspace $\ZS_c$.

The GIN method of \textcite{sorrenson2019disentanglement} achieves this by defining $p(\Z)$ as a mixture of diagonal Gaussian distributions.
Moreover, the authors prove theoretically that this method induces a code disentanglement under certain conditions.
For the MNIST dataset with $\dim(\x)=784$, the mixture naturally consists of one component per class label $y(\x)\in\{0,...,9\}$ with learnable means $\mub(y(\x))$ and variances $\sigmab^2(y(\x))$.
Maximum-likelihood training of an incompressible flow $g$ by formula (\ref{eq:incompressible-cov}) converges to core codes with $\dim(\z_c)=22$, and the remaining dimensions $\z_d$ can be set to zero without visible effects on the generated images.
Thus, the manifolds defined by $g\big([\z_c;\; \0]^T\big)$, i.e. by the core codes padded with zeros, completely cover the MNIST handwriting style variability.

{\em Nested dropout normalizing flows} \autocite{bekasovs2020ordering} enforce an importance-ordering of the code dimensions by applying dropout to them:
In each training iteration, an index $k$ is sampled from a geometric distribution over the integers $1...C$, and all code dimensions $\z_{>k}$ are set to zero before decoding.
The squared error of the resulting reconstruction is used as a training objective, as in $\MS$-flows.
But unlike $\MS$-flows, the set of core dimensions is not fixed beforehand here. 
Instead, dimensions that get dropped out rarely (i.e. those with small index) will learn to represent the crucial information, leading naturally to the desired ordering.
Moreover, \textcite{bekasovs2020ordering} combine the dropout-based reconstruction objective with the usual maximum likelihood objective, so that the model also learns to generate the full data distribution in the absence of dropout.
Experiments clearly show an increase in generated detail as the number of active code dimension increases, but it is as yet unclear to what degree this ordering corresponds to a semantic disentanglement, as in GIN.

{\em Independent mechanism analysis} \autocite{gresele2021independent} and {\em principal component flows} \autocite{cunningham2022principal} enforce disentanglement by the constraint that the flow's Jacobians should have orthogonal rows for every $\x$, i.e. $\J_f\!\cdot\!\J_f^T$ is diagonal everywhere.%
\footnote{Note that this condition is much weaker than the requirement for conformal maps -- there, all diagonal elements of $\J_f\!\cdot\!\J_f^T$ must be equal, cf. \autoref{sec:data-on-embedded-manifolds}. Moreover, in practice the constraint can only be checked or enforced at the training points and may be invalid elsewhere.} 
It implies that axis-aligned changes in code space will cause orthogonal changes in data space.
In other words, the Jacobians span local coordinate systems at each point $\x$ similar to PCA, but adapted to the local geometry of $p(\X)$ in a neighborhood of $\x$ -- while PCA coordinate systems are globally constant, disentangled coordinate systems may be rotated and scaled differently at every point.
The Jacobian determinant then simplifies into
\begin{equation}\label{eq:disentangled-jacobian}
    \big|\det\big(\J_f(\x)\big)\big| \overset{!}{=}\prod_{j=1}^D \big(\J_f(\x)_j\cdot \J_f(\x)^T_j\big)^{1/2}
    = \prod_{j=1}^D \big\|\J_f(\x)_j\big\|_2
\end{equation}
with $\J_f(\x)_j$ denoting row $j$.
The effect of the constraint can be best understood by a hierarchical decomposition of the code space.
Let $\TS$ be a binary tree, whose leaves $L_j$ correspond to individual code dimensions $\Z_j$, and whose interior nodes $N_k$ represent the union of the subspaces of their left and right child, $\ZS_k=\ZS_{\text{left}(k)}\cup\ZS_{\text{right}(k)}$. 
The root node contains the entire code space $\ZS$.

Each node defines a family of manifolds $\MS_k(\z_{\setminus k})\subset\XS$ and associated random variables $\X_k(\z_{\setminus k})=g(\Z_k; \z_{\setminus k})$, which are obtained by varying the coordinates in $\ZS_k$ and keeping the other coordinates $\z_{\setminus k}\in\ZS\setminus\ZS_k$ fixed.
Assuming that the variables in $\ZS_k$ are distributed according to $p(\Z_k)$, one can use the autoencoder CoV formula (\ref{eq:autoencoder-cov}) to define a density $p\big(\X_k(\z_{\setminus k})\big)$ within each of these manifolds.
The authors of \autocite{cunningham2022principal} use this to define the {\em pointwise mutual information} as
\begin{equation}
    \IS(\z_k; \z_{\setminus k}) = \log \frac{p\big(\X_k\!=\!g(\z_k; \z_{\setminus k})\big)}{p\big(\X_{\text{left}(k)}\!=\!g(\z_{\text{left}(k)}; \z_{\setminus k\,\cup\,\text{right}(k)})\big)\cdot p\big(\X_{\text{right}(k)}\!=\!g(\z_{\text{right}(k)}; \z_{\setminus k\,\cup\,\text{left}(k)})\big)}
\end{equation}
It measures by how much the information content of the manifold density $p(\X_k)$ exceeds the information content of the product density of its children $p(\X_{\text{left}(k)})\cdot p(\X_{\text{right}(k)})$, or how much information arises from the children's entanglement.
This allows the decomposition of the bijective CoV formula (\ref{eq:bijective-cov}) into
\begin{align}
    \hspace{-5mm}\CoV{hierarchical decomposition CoV} \nonumber \\
    p(\X\!=\!\x) =  
    \bigg(\prod_{j=1}^D p\big(\Z_j\!=f_j(&\x)\big) \cdot \big\|\J_f(\x)_j\big\|_2 \bigg)
    \cdot \exp\Big(\sum_{k\in\TS} \IS\big(\z_k\!=\!f_k(\x);\, \z_{\setminus k}\!=\!f_{\setminus k}(\x)\big)\Big)
        \label{eq:hierarchical-decomposition-cov}
\end{align}
The first factor represents the change-of-variables contribution of the individual dimensions of $\Z$, i.e. of the leaves of $\TS$, and the second one the interaction between these dimensions according to the remaining nodes of $\TS$.
Note that this formula is valid for any tree decomposition $\TS$.
Disentanglement is equivalent to the requirement that the interactions vanish, so that $\exp\big(\sum_k \IS\big(\z_k\!=\!f_k(\x); \z_{\setminus k}\!=\!f_{\setminus k}(\x)\big)\big)=1$ in every point $\x$.
The cited papers 
achieve this by enforcing equality in (\ref{eq:disentangled-jacobian}) as an additional regularizer in the training objective.
They then define the core subspace by the property that the norms of the corresponding rows of the Jacobian exceed a threshold, $\ZS_c=\{\Z_j: \E_{p^*(\x)}\big[\|\J_f(\x)_j\|_2\big] \ge \epsilon\}$, similar to variable selection by linear PCA.

A promising alternative was recently proposed by \textcite{horvat2022intrinsic}. 
The authors observe that latent dimensions in the core and detail subspaces exhibit quite distinct behavior under additive noise augmentation of the data.%
\footnote{Since they do not assume disentanglement, they analyze these phenomena for the singular values of the Jacobian, but the argument applies equivalently to the Jacobian row norms of a disentangled representation.}
The Jacobian row norms of detail dimensions tend to be proportional to the strength of the added noise (reflecting a corresponding widening of the data manifold due to noise components perpendicular to it), whereas the norms of core dimensions remain approximately constant (indicating that noise components within the manifold have little effect as long as the noise is not too strong). 
Thus, the latent dimensions can be classified according to the stability of the respective row norms under increasing amounts of noise, and this criterion seems to be very robust across datasets.

Either criterion finally results in a CoV formula that is a product of 1-dimensional contributions
\begin{align}
    \hspace{-5mm}\CoV{disentangled flow CoV} \nonumber \\
    p(\X\!=\!\x) =  
    \bigg(\prod_{j=1}^C &p\big(\Z_j\!=\!f_j(\x)\big) \cdot \big\|\J_f(\x)_j\big\|_2 \bigg) \cdot
    \bigg(\prod_{j=C+1}^D p\big(\Z_j\!=\!f(\x)_j\big) \cdot \big\|\J_f(\x)_j\big\|_2 \bigg)
        \label{eq:disentangled-flow-cov}
\end{align}
The first product refers to the manifold $\MS_c$ spanned by the core dimensions, and the second one to the fibers $\FS(\z_c)$ defined by the detail dimensions for fixed $\z_c$.
Thanks to (\ref{eq:disentangled-jacobian}), the CoV formulas (\ref{eq:bijective-cov}) and (\ref{eq:disentangled-flow-cov}) are equivalent in disentangled flows, and one can base calculations on the cheaper variant for the particular application.
However, it remains an open problem how well this method scales to higher dimensions and under which conditions good disentanglements are identifiable \autocite{hyvarinen2023nonlinear}.

\section{Connections to Coding Theory}
\label{sec:coding-theory}

Encoder-decoder architectures are the classical subject of {\em coding theory} or the theory of compression \autocite{cover2006elements}.
It deals with the design of codes that can be transmitted efficiently over a communication channel of finite capacity.
Classical coding theory as developed by Shannon  considers the trade-off between {\em rate} (the average amount of compression) and {\em distortion} (the expected reconstruction error).
However, recent research from the perspective of generative modeling revealed that the trade-off actually involves a third component: the distribution shift between the original and decoded data \autocite{blau2019rethinking}.
The third component is also called ``perceptual quality'', because it measures the perceived realism of the reconstructed data, i.e. the faithful representation of details and the absence of artifacts.
Perceptual quality is, for example, measured and encouraged by the discriminator of a GAN \autocite{goodfellow2014generative,agustsson2023multi}, whereas low distortion is optimized by the reconstruction error of an autoencoder.

To illustrate the relevance of these concepts for our categorization of generative models, we first recall some important definitions. 
An encoder-decoder model involves three random variables $\X$ (the original data), $\Z$ (the codes), and $\hat{\X}$ (the reconstructed data), whose joint distribution is defined as
\begin{equation}
    p(\X, \Z, \hat{\X}) = p^*(\X)\cdot p(\Z\given\X) \cdot p(\hat{\X}\given\Z)
\end{equation}
where $p^*(\X)$ is the true data distribution, $p(\Z\given\X)$ the encoder, and $p(\hat{\X}\given\Z)$ the decoder.
When encoder and/or decoder are deterministic, the corresponding conditionals reduce to delta distributions.
The conditional distribution of the reconstructions, given an original data instance $\x$, and the generative distribution are obtained by marginalization:
\begin{align}
    p(\hat{\X}\given\X\!=\!\x) =& \int p(\Z\!=\!\z\given\X\!=\!\x) \cdot p(\hat{\X}\given\Z\!=\!\z)\,d\z \\
    p(\hat{\X}) =& \int p^*(\X\!=\!\x) \cdot p(\hat{\X}\given\X\!=\!\x)\, d\x
\end{align}
For some distance $\delta$ (e.g. Hamming, squared, or perceptual), the distortion measures the expected reconstruction error for individual instances 
\begin{equation}
    \bar\delta = \E_{\x,\widehat{\x}\sim p^*(\X)\cdot p(\hat{\X}\given\X)}\big[\delta(\x, \widehat{\x})\big]
\end{equation}
Similarly, for some divergence $\Delta$ (e.g. KL, total variation, Wasserstein), the perceptual quality index measures the distribution shift between the true and generated data
\begin{equation}
    \bar\Delta = \Delta\big[p^*(\X) \,||\; p(\hat{\X})\big]
\end{equation}
\textcite{blau2019rethinking} now define the {\em rate-distortion-perception function} as
\begin{equation}\label{eq:rate-distortion-perception-function}
    R(\delta_\text{max}, \Delta_\text{max}) = \min_{p(\hat{\X}\given\X)}\IS[\X, \hat{\X}]\qquad\text{s.t.}\qquad \bar\delta\le\delta_\text{max}\quad\text{and}\quad \bar\Delta\le\Delta_\text{max}
\end{equation}
where $\IS$ is the mutual information between original and reconstructed data, and the minimum is taken over all possible encoders and decoders from our chosen family.
The mutual information effectively measures the capacity of the codes $\Z$ and thus depends on the bottleneck size.
The classical rate-distortion function is recovered by setting $\Delta_\text{max}=\infty$, i.e. by ignoring the perceptual quality index.

\begin{figure}[t]
    \centering
    \includegraphics[width=.6\linewidth]{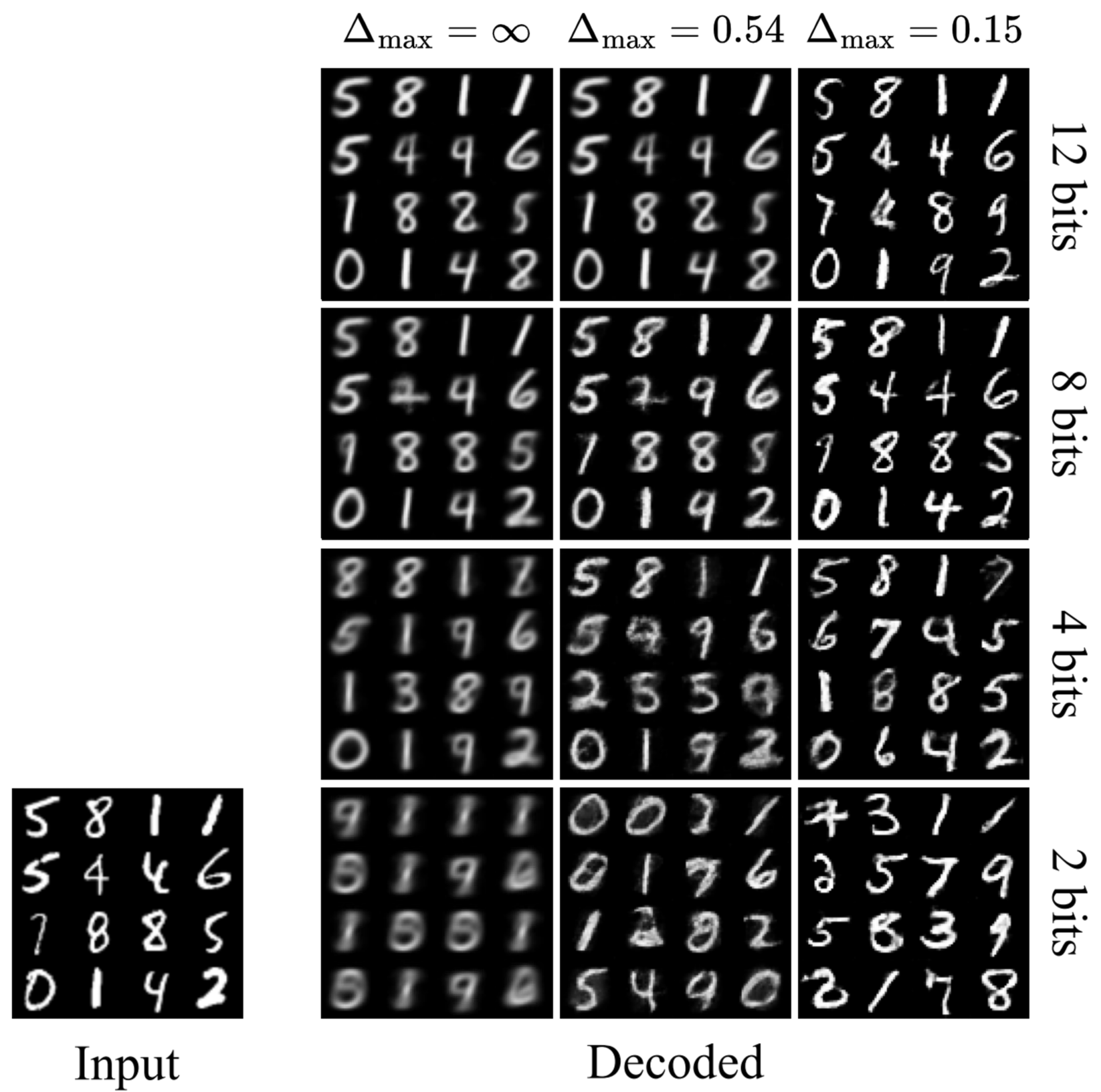}
    \caption{Effect of the rate-distortion-perception trade-off in distribution-preserving lossy compression (see \autoref{sec:finite-autoencoders}) for MNIST: 
    Training minimizes the distortion for a given code capacity (rows) and perception quality bound (columns). 
    When perception quality is ignored ($\Delta_\text{max}=\infty$), the optimal reconstructions are unrealistic and cartoon-like. 
    High emphasis on perception quality ($\Delta_\text{max}=0.15$) leads to realistic diversity of the reconstructed images at the price of increased reconstruction error. 
    When the rate is too low (4 bits or less), not even the digit labels are preserved. Figure from \autocite{blau2019rethinking}.}
    \label{fig:mnist-rate-distortion-perception-trade-off}
\end{figure}
The crucial insight of \textcite{blau2019rethinking} is a proof that the simultaneous minimization of all three terms in (\ref{eq:rate-distortion-perception-function}) is impossible.
For example, minimizing distortion at a fixed rate leads to simplified, cartoon-like reconstructions that lack realism and are easily recognizable as synthetic, see \autoref{fig:mnist-rate-distortion-perception-trade-off}.
Fortunately, the trade-off is rather mild: the conflict between distortion and realism typically manifests only at relatively low rates.
Moreover, \textcite{blau2019rethinking} proved for the squared reconstruction error, $\delta(\x, \widehat{\x})=\norm{\x - \widehat{\x}}^2$, that perfect realism at a fixed rate is always achievable with no more than a two-fold increase in distortion.
Our four types of generative models represent different resolutions of the ensuing triple trade-off:
\begin{itemize}
    \item Bijective flows achieve zero distortion and zero distribution shift at the price of lossless encoding.
    \item Injective flows allow to adjust the compression rate while maintaining a low reconstruction error, but sacrifice the diversity of the reconstructed data: $p(\hat{\X})$ is restricted to the decoder manifold $\MS$ and does not cover the entire data domain $\XS$.
    \item Stochastic flows generate the correct data distribution $p(\hat{\X})\approx p^*(\X)$ at any compression rate, but result in non-deterministic codes $\z\sim p(\Z\given\X\!=\!\x)$. 
    This causes an increase in reconstruction error.
    It is even possible \autocite{chen2017variational,zhao2019infovae} that the decoder ignores the codes and acts as an unconditional generative model for the target distribution, in which case inputs and reconstructions are unrelated and the expected reconstruction error assumes its maximum.
    \item Split flows allow free adjustment of the trade-off between all three objectives.
    An initial, albeit heavily engineered, architecture with a tunable adjustment between realism and distortion is proposed in \autocite{agustsson2023multi}.
\end{itemize}









\section{Efficient Computation of Jacobian Determinants}\label{sec:efficient-jacobians}

Change-of-variables formulas are only useful in practice when the Jacobian determinants that adjust the probability for local volume contraction or expansion can be computed efficiently.
The simplest special case are {\em incompressible flows}, which are designed such that $\det(\J_f)=\det(\J_g)=1$ for all $\x$ and $\z$ and thus $p(\X=\x)=p(\Z=f(\x))$.
Another simple case is a linear decoder/encoder pair
\begin{equation}\label{eq:linear-normalizing-flow}
    g(\z) = \W \cdot \z  + \w; \qquad\qquad f(\x) = \W^{-1} \cdot (\x - \w) 
\end{equation}
Its Jacobians are independent of the current data points $\x$ resp. $\z$
\begin{equation}
    \J_f = \W^{-1}; \qquad\qquad \J_g = \W
\end{equation}
and the determinants for the CoV formula can be precomputed by standard linear algebra methods even for high dimensions.
However, when $p(\Z) = \normal(0, \eye)$, this model can only generate Gaussian distributions $p(\X)$.
If we nonetheless try to model a non-Gaussian $p^*(\X)$, we can choose $\W$ (by PCA) such that the encoder pushforward distribution $p_E(\Z) = f_\#p^*(\X)$ has the desired unit covariance, but non-zero higher moments will inevitably remain.
Thus, $p_E(\Z)$ is still a complicated distribution, and $p_E(\Z) = p(\Z)$ is in general unachievable with a standard normal $p(\Z)$ or other predefined simple choices.

When $f(\z)$ and $g(\z)$ are non-linear, $\J_f(\x)$ resp. $\J_g(\z)$ must be recomputed for every data point.
While this is in principle easy to do with modern auto-differentiation libraries, it is only computationally feasible for low and medium dimensions, especially when it must be repeated for every training iteration over many epochs.
The effort can be drastically reduced when the Jacobians have special structure.
A common trick is to define $g(\x)$ separately for each element of $\x$.
To this end, the target distribution $p(\X)$ is factorized according to the Bayesian chain rule
\begin{equation}\label{eq:bayesian-chain-rule}
    p(\X) = p_1(X_1)\ \prod_{j=2}^{D} p_j(X_j\given\X_{1:j-1})
\end{equation}
and the decoder defines a one-dimensional pushforward for every conditional 
\begin{align}
    x_1 \sim p_1(X_1) \qquad\Longleftrightarrow\qquad & x_1 = g_1(z_1) \ \text{ with }\ z_1 \sim p_1(Z_1) \nonumber \\
    x_j \sim p_j(X_j\given\X_{1:j-1}) \qquad\Longleftrightarrow\qquad & x_j = g_j(z_j;\ \x_{1:j-1}) \ \text{ with }\ z_j \sim p_j(Z_j)
\end{align}
such that the functions $g_j(z_j;\ \x_{1:j-1})$ are invertible in $z_j$.
This approach is known as a {\em Knothe-Rosenblatt rearrangement} (see \autocite{knothe1957contributions,rosenblatt1952remarks} for the original ideas and \autocite{marzouk2016sampling} for a modern treatment) or an {\em auto-regressive flow} (e.g. \autocite{papamakarios2017masked,huang2018neural}).
When $p(\X)$ has a non-degenerate density, such a rearrangement always exists regardless of the index permutation used in the decomposition (\ref{eq:bayesian-chain-rule}).
Crucially, the decomposition ensures that $\J_g$ is a triangular matrix, so that its determinant reduces to the product of diagonal elements and becomes cheaply computable:
\begin{equation}\label{eq:triangular-jacobian}
    \det\big(\J_g(\z)\big) = \prod_{j=1}^D \big(\J_g(\z)\big)_{jj} = \prod_{j=1}^D \left.\frac{\partial g_j(z_j;\ \x_{1:j-1})}{\partial z_j}\right|_{z_j}
\end{equation}
An index permutation of special interest is the {\em causal decomposition}, where each $p_j$ is conditioned on the (possibly empty) set of causal parents $\text{PA}(X_j)$ of the variable $X_j$, as specified by an acyclic causal graph.
Then, the decoder is known as a {\em structural causal model} (SCM) and takes the form
\begin{equation}
    x_j \sim p_j(X_j\given\text{PA}(X_j)) \qquad\Longleftrightarrow\qquad x_j = g_j(z_j;\ \text{PA}(x_j)) \ \text{ with }\ z_j \sim p_j(Z_j)
\end{equation}
The triangular determinant formula (\ref{eq:triangular-jacobian}) applies likewise. 
The functions $g_j$ are simpler for the causal decomposition than for alternative index orders, and learning of SCMs is a hot topic in causal inference, e.g. \autocite{shimizu2014lingam,peters2014causal,xia2021causal},
and the connection to autoregressive flows is explicitly explored in \autocite{khemakhem2021causal}.

A currently very popular architecture, which appears to be easier to learn, decomposes $g(\z)$ into a sequence of $L$ simpler bijective layers:
\begin{equation}
    \x = g(\z) = (g_L \circ ... \circ g_1)(\z)
\end{equation}
with intermediate variables defined by $\z^{(l)}=g_l\big(\z^{(l-1)}\big)$ and $\z^{(0)} = \z$, $\x = \z^{(L)}$. 
The determinant of $\J_g$ is then the product of the determinants of the layers' Jacobians
\begin{equation}
    \det\big(\J_g(\z)\big) = \prod_{l=1}^L \det\big(\J_{g_l}\big(\z^{(l-1)}\big)\big) = \prod_{l=1}^L \det\left(\left.\frac{\partial g_l\big(\z^{(l-1)}\big)}{\partial \z^{(l-1)}}\right|_{\z^{(l-1)}}\right)
\end{equation}
To ensure efficient computation of the individual determinants, the layer functions $g_l$ are alternatingly selected among three possibilities:
\begin{itemize}
    \item orthogonal layers: $g_l$ is linear (equation (\ref{eq:linear-normalizing-flow})) with $\W$ orthogonal, such that $\det(\W) = 1$
    \item actnorm layers: $g_l$ is elementwise linear, i.e. $\W$ is diagonal with $\det(\W) = \prod_j \W_{jj}$
    \item coupling layers: $g_l$ implements an affine coupling transformation
    \begin{align}\label{eq:affine-coupling}
        \z^{(l)}_{1:D'} = &\, \z^{(l-1)}_{1:D'} \nonumber \\
        \z^{(l)}_{D'+1:D} = &\, \diag\left(\s^{(l)}\big(\z^{(l-1)}_{1:D'}\big)\right) \cdot \z^{(l-1)}_{D'+1:D} + \w^{(l)}\big(\z^{(l-1)}_{1:D'}\big)
    \end{align}
    with $D'<D$ a hyperparameter (usually $D'=D/2$) and $\s^{(l)}>0$ and $\w^{(l)}$ learnable vector functions of output dimension $D-D'$. 
    A coupling layer obviously has a triangular Jacobian $\J_{g_l}$, and the determinant is 
    \begin{equation}
        \det\left(\J_{g_l}\big(\z^{(l-1)}\big)\right) = \prod_{j=D'+1}^D \s^{(l)}\big(\z^{(l-1)}_{1:D'}\big)_j
    \end{equation}
\end{itemize}
This design is known as the RealNVP \autocite{dinh2017density} and GLOW \autocite{kingma2018glow}. 
Improved variants replace the affine transformation in (\ref{eq:affine-coupling}) with more complex invertible functions, e.g. quadratic or cubic B-splines \autocite{muller2019neural,durkan2019cubic-spline}, linear-rational or rational-quadratic splines \autocite{dolatabadi2020invertible,durkan2019neural} or sum-of-squares polynomials \autocite{jaini2019sum}, while maintaining the triangular form of $\J_{g_l}$ and therefore the efficient computation of the determinant.

In the absence of special structure, efficient computation of the Jacobian determinant is much harder and a hot research topic.
Many ideas build on the identity 
\begin{equation}
    \log(\det(\J)) = \tr(\log(\J))
\end{equation}
for non-singular square $\J$.
The matrix $\log(\J)$ is now approximated by some matrix expression $\A$, e.g. using Chebyshev polynomials \autocite{han2015large}, truncated power series \autocite{boutsidis2017randomized,chen2019residual}, or Riemann–Stieltjes integrals \autocite{ubaru2017fast}.
Finally, the trace of $\A$ is estimated by randomization 
\begin{equation}
    \tr(\A)=\E_\epsilon[\epsilon^T \A\, \epsilon] \approx \frac{1}{M}\sum_{m=1}^M \epsilon_m^T \A\, \epsilon_m\qquad
\end{equation}
with $\epsilon\!\sim\!p(\epsilon)$ and $\E[\epsilon\cdot\epsilon^T] = \eye$.
This is known as Hutchinson's trace estimator \autocite{hutchinson1989stochastic}.
The crucial benefit over computing $\tr(\log(\J))$ is that the resulting expressions $\epsilon_m^T \A\, \epsilon_m$ can be calculated solely in terms of Jacobian-vector and vector-Jacobian products, which modern auto-differentiation libraries implement efficiently without ever constructing $\J$ or $\log(\J)$.
These techniques are mostly suitable for inference on a trained model and add significant computational overhead if used during maximum-likelihood learning.

Since learning via gradient descent requires the {\em derivative} of the log determinant with respect to the learnable parameters $\theta$, it is more efficient to approximate $\nabla_\theta \log(\det(\J))$  directly during training.
If the model is restricted to be fully connected without residual connections, this derivative can be cheaply obtained by replacing additive parameter updates with multiplicative ones \autocite{gresele2020relative} or by simultaneously learning the inverse of the weight matrices \autocite{keller2021self}. 
Extensions of these methods to convolutional networks are also possible.

For unconstrained injective flows implementing the autoencoder CoV (\ref{eq:autoencoder-cov}), \autocite{caterini2021rectangular} propose an analog to the above stochastic trace-estimator for the derivative.
It is based on the identity
\begin{equation}
    \frac{\partial}{\partial \theta} \frac{1}{2} \log \det \big(\J_g^T\cdot\J_g\big) = \frac{1}{2} \tr \left( \big(\J_g^T\cdot\J_g\big)^{-1} \frac{\partial}{\partial \theta} \big(\J_g^T\cdot\J_g\big)\right) 
\end{equation}
where $\theta$ are the decoder's learnable parameters.
The trace can be approximated by Monte-Carlo sampling: 
\begin{equation}
    \frac{1}{2} \tr \left( \big(\J_g^T\cdot\J_g\big)^{-1} \frac{\partial}{\partial \theta} \big(\J_g^T\cdot\J_g\big)\right) \approx \frac{1}{2M} \sum_{m=1}^M \epsilon_m^T \big(\J_g^T\cdot\J_g\big)^{-1} \frac{\partial}{\partial \theta} \big(\J_g^T\cdot\J_g\big) \epsilon_m
\end{equation}
where again $\epsilon\!\sim\!p(\epsilon)$ with $\E[\epsilon\cdot\epsilon^T] = \eye$.
A drawback of this formulation is that the term $\epsilon_m^T \big(\J_g^T\cdot\J_g\big)^{-1}$ must be evaluated by an iterative linear solver (e.g. conjugate gradients) whose complexity scales linearly with the dimension of the code space.
An alternative trace estimator avoiding this problem was recently proposed in \autocite{sorrenson2023maximum}, who noticed that the pseudo-inverse of $\J_g$ equals the Jacobian $\J_f$ of the encoder.
This results in the estimator
\begin{equation}
    \frac{1}{2} \tr \left( \big(\J_g^T\cdot\J_g\big)^{-1} \frac{\partial}{\partial \theta} \big(\J_g^T\cdot\J_g\big)\right) = \tr\left( \J_{\!f}\, \frac{\partial}{\partial \theta} \J_g \right) \approx \frac{1}{M} \sum_{m=1}^M \epsilon_m^T \J_{\!f}\,\, \frac{\partial}{\partial \theta} \J_g\ \epsilon_m
\end{equation}
It turns out that $M=1$ is sufficient for this estimator to be effective in the context of stochastic optimization, so that processing of a training batch in an injective flow takes only about twice as long as in a corresponding traditional autoencoder, which is trained by reconstruction error alone.


\section{Acknowledgments}

This work was supported by Deutsche Forschungsgemeinschaft (DFG, German Research Foundation) under Germany’s Excellence Strategy EXC-2181/1 - 390900948 (the Heidelberg STRUCTURES Cluster of Excellence), Informatics for Life funded by the Klaus Tschira Foundation, and Model-based AI funded by the Carl-Zeiss-Stiftung.

\appendix

\section{Analytic Treatment of the Donut Distribution}\label{sec:donut-distribution}

In this section, we will consider a simple toy distribution to demonstrate the differences between the fundamental types of generative models we discussed above.
Consider a 2-dimensional uniform distribution over a donut with inner (hole) radius $R_0=3$ and outer radius $R_1=8$, see \autoref{fig:donut}.
By construction, the true density on the donut is the inverse of the area
\begin{equation}
    p^*(\X) = \frac{1}{\pi(R^2_1 - R^2_0)}
\end{equation}
The simplest representation of data from the above distribution is obtained by a deterministic autoencoder with a 1-dimensional bottleneck.
When trained with squared reconstruction loss, it learns the 1-D manifold $\MS$ with minimal average distance from the data points.
Thanks to the symmetry of our toy problem, this manifold is a circle with radius $R_\MS = \frac{2}{3} (R^3_1 - R^3_0) / (R^2_1 - R^2_0) = \frac{194}{33}\approx 5.88$.
The encoder thus calculates the polar angle of a given data point and ignores its radial position.
The induced code distribution is uniform:
\begin{equation}\label{eq:donut-autoencoder}
    z = f(\x) := \arg(\x)\qquad\Rightarrow\qquad p(Z) = \uniform(0,\,2\pi) = \frac{1}{2\pi}
\end{equation}
The fibers $\FS(z)$ of this transformation are the radial lines with constant angle $\arg(\x)$.
The decoder generates data uniformly on the circle using the generative model
\begin{equation}
    \widehat{\x} = g(z) := 
    \begin{pmatrix}
       R_\MS\cdot\cos(z)\\
       R_\MS\cdot\sin(z)
    \end{pmatrix}
    \qquad\text{with}\quad
    z\sim\uniform(0,\, 2\pi)
\end{equation}
The decoder's Jacobian is $\J_g=(-R_\MS\cdot\sin(z),\,R_\MS\cdot\cos(z))^T$, so that $|\det(\J^T_g\cdot \J_g)|^{-1/2} = \frac{1}{R_\MS}$.
Inserting this into the autoencoder change-of-variables formula (\ref{eq:autoencoder-cov}) gives
\begin{equation}
    p(\widehat{\x}\in\MS) = \frac{1}{R_\MS}\cdot\uniform(0,\, 2\pi) = \frac{1}{2\pi R_\MS}
\end{equation}
This model has only one degree of freedom, namely $z$, and the joint distribution of data and codes is $p(\widehat{\x},z)=p(z)\cdot \delta\big(\widehat{\x}^T -(R_\MS\cdot\cos(z),\,R_\MS\cdot\sin(z))\big)$.

A split flow preserves the deterministic encoder (\ref{eq:donut-autoencoder}) and extends the decoder so that it generates the full data distribution.
We achieve this by an additional conditional density $p(\X\in\FS(z)\given \Z\!=\!z)$ that models the data behavior within the fiber $\FS(z)$.
The resulting density is expressed by the change-of-variables formula (\ref{eq:split-nf-cov}).
Representing $\x$ in polar coordinates $\x=(r ,\alpha)$, the conditional simplifies into $p(R)$ independent of $Z$, since the radial behavior of the fibers does not change for different angles $z=\alpha$.
The Jacobian determinant of the transformation from Cartesian to polar coordinates is $1/r$, and we arrive at the change-of-variables formula
\begin{equation}\label{eq:donut-stochastic-nullspace-cov-raw}
    p(\X\!=\!\x)=p\big(Z=\arg(\x)\big)\cdot p\big(R=\norm{\x}\big)\cdot\frac{1}{\norm{\x}}
\end{equation}
By definition, the LHS is a constant (the uniform distribution on the donut), so the distribution $p(R)$ must be proportional to $r=\norm{\x}$ to cancel out the Jacobian term.
After normalization of $p(R)$ in the interval $(R_0,\, R_1)$, we arrive at the generative model
\begin{equation}\label{eq:donut-stochastic-nullspace-decoder}
    \x = g(z; r) := 
    \begin{pmatrix}
       r\cdot \cos(z)\\
       r\cdot\sin(z)
    \end{pmatrix} 
    \;\;\text{with}\;\;
    \begin{cases}
        z \sim \uniform(0,\,2\pi) \\
        r \sim p(R),\quad p(R\!=\!r) = 2r/(R^2_1-R^2_0),\;\;r\in(R_0,\,R_1)
    \end{cases}
\end{equation}
Inserting these formulas into equation (\ref{eq:donut-stochastic-nullspace-cov-raw}) gives the model density as
\begin{equation}
    p(\X\!=\!\x)=\frac{1}{2\pi}\cdot \frac{2\norm{\x}}{(R^2_1 - R^2_0)} \cdot \frac{1}{\norm{\x}} = \frac{1}{\pi(R^2_1 - R^2_0)} = p^*(\X)
\end{equation}
which happens to be exact for our simple toy problem.
This model has two degrees of freedom $z$ and $r$, and the joint distribution of data and codes is $p(\x, z)=p(z)\cdot p(r)\cdot \delta\big(\x^T -(r\cos(z),\,r\sin(z))\big)$.
The decoder can also be interpreted as a split normalizing flow with codes $\z = (z,r)$, where the component $\z_1=z$ represents the core property of the data (the polar angle of $\x$ on the circle $\MS$) and $\z_2=r$ the details (the radial position of $\x$).

A standard normalizing flow similarly maps the data to 2-dimensional codes, but does not enforce a split into core and detail dimensions and instead uses a standard normal code distribution
\begin{equation}
    p(\Z\!=\!\z)=\normal(0,\eye_2)=\frac{1}{2\pi}\exp(-\norm{\z}^2/2)
\end{equation}
To derive an analytic generative model, we first transform the normally distributed $\z$ into two independent uniformly distributed variables by the inverse Box-Muller transform:
\begin{alignat}{3}
    \alpha =&\, \arg(\z)\qquad &&\Rightarrow \qquad \alpha &&\sim\uniform(0,\,2\pi) \\
    \rho = &\, 1 - \exp(-\norm{\z}^2/2)\qquad &&\Rightarrow \qquad \rho &&\sim\uniform(0,\,1)
\end{alignat}
Building upon equation (\ref{eq:donut-stochastic-nullspace-decoder}), the variable $\rho\sim\uniform(0,\,1)$ is now transformed to $p(r) = 2r / (R^2_1-R^2_0)$ by the inverse probability integral transform $r=(\rho\cdot(R^2_1-R^2_0) + R^2_0)^{1/2}$.
The resulting generative model becomes
\begin{equation}\label{eq:donut-normalizing-flow}
    \x = g(\z) := 
    \begin{pmatrix}
       r\cdot \cos(\alpha)\\
       r\cdot\sin(\alpha)
    \end{pmatrix} 
    \quad\text{with}\quad
    \begin{cases}
        \z\sim\normal(0, \eye_2) \\
        \alpha = \arg(\z) \\
        r = \big((1-\exp(-\norm{\z}^2/2))
\cdot(R^2_1-R^2_0) + R^2_0\big)^{1/2}
    \end{cases}
\end{equation}
A lengthy calculation shows that the Jacobian determinant of the normalizing flow's generative model is $\det(\J_g) = \exp(-\norm{\z}^2/2)(R^2_1-R^2_0)/2$.
Inserting this into the change-of-variables formula (\ref{eq:bijective-cov}) gives the model density
\begin{align}
    p(\X\!=\!g(\z)) =&\, p(\Z\!=\!\z)\cdot |\det(\J_g)|^{-1} \nonumber \\
     =&\, \frac{1}{2\pi}\exp(-\norm{\z}^2/2)\cdot \frac{2}{\exp(-\norm{\z}^2/2)(R^2_1-R^2_0)} \\
     =&\, \frac{1}{\pi(R^2_1 - R^2_0)} = p^*(\X) \nonumber
\end{align}
which is again equal to the true density.
This model also has two degrees of freedom, and the joint distribution of data and codes is $p(\x, \z)=p(\z)\cdot \delta\big(\x^T -(r\cos(\alpha),\,r\sin(\alpha))\big)$ with $r$ and $\alpha$ calculated from $\z$ according to equation (\ref{eq:donut-normalizing-flow}).

Finally, a variational autoencoder learns a full joint distribution of both $\x$ and $z$, where we take $z$ to be 1-dimensional.
In contrast to the plain autoencoder, the mapping from $\x$ to $z$ is now non-deterministic, and there is no unique code for a given $\x$.
For the sake of illustration, let us define a probabilistic encoder that maps each data point uniformly to a set of codes in an interval around the true polar angle $\alpha = \arg(\x)$:
\begin{equation}
    p(Z\given\X\!=\!\x) = \uniform\big(\arg(\x)-\alpha_0,\, \arg(\x)+\alpha_0\big)
\end{equation}
where $\alpha_0$ is the interval's half width (say, $\alpha_0=5^\circ$).
The induced marginal code distribution is $p(Z) = \E_{\x\sim p(\x)}\big[p(Z\given \X\!=\!\x)\big]=\uniform(0,\, 2\pi)$.
The corresponding decoder maps a given code $z$ uniformly to a {\em segment} of the donut between the angles $z-\alpha_0$ and $z+\alpha_0$. 
The resulting generative model is similar to equation (\ref{eq:donut-stochastic-nullspace-decoder}), but instead of sampling the angle uniformly from $(0,2\pi)$, one samples it only in the $\alpha_0$ interval around $z$:
\begin{equation}
    \x \sim p(\X\given Z\!=\!z)\;\;\Longleftrightarrow\;\;
    \x = 
    \begin{pmatrix}
       r\cdot \cos(\alpha)\\
       r\cdot\sin(\alpha)
    \end{pmatrix} 
    \;\;\text{with}\;\;
    \begin{cases}
        \alpha \sim \uniform(z-\alpha_0,\,z+\alpha_0) \\
        r \sim p(R),\;\; p(R\!=\!r)=2r/(R^2_1-R^2_0)
    \end{cases}
\end{equation}
The data distribution is recovered by marginalization $p(\X) = \E_{z\sim p(Z)}\big[p(\X\given Z=z)\big]$.
It can be easily checked that this model is self-consistent. \todo{(do it?)}
The joint distribution $p(\X,Z)=p(\X)\cdot p(Z\given \X) = p(Z)\cdot p(\X\given Z)$ has three degrees of freedom, because the 2-dimensional random variable $\X$ and the 1-dimensional $Z$ are no longer deterministically related.

\clearpage
\FloatBarrier
\printbibliography

\end{document}